\documentclass[sn-mathphys]{sn-jnl2}%

\usepackage{times}
\usepackage{epsfig}
\usepackage{graphicx}
\usepackage{amsmath}
\usepackage{amssymb}
\usepackage{color}
\usepackage{nicefrac}
\usepackage{siunitx} %
\usepackage[mathscr]{euscript}
\usepackage{microtype}
\usepackage{booktabs}
\usepackage{wrapfig}
\usepackage{caption}
\usepackage{subcaption}
\usepackage[inline]{enumitem}
\usepackage{float}

\jyear{2022}%

\theoremstyle{thmstyleone}%

\theoremstyle{thmstyletwo}%

\theoremstyle{thmstylethree}%

\renewcommand{\R}{\mathbb{R}}

\newif\ifnotes
\notestrue %
\ifnotes
\newcommand{\snote}[1]{\textcolor{red}{\textbf{Skylar}:#1}}
\newcommand{\bnote}[1]{\textcolor{green}{\textbf{Bernie}:#1}}
\else
\newcommand{\snote}[1]{}
\newcommand{\bnote}[1]{}
\fi

\begin{document}

\title{Building 3D Generative Models from Minimal Data}

\author*[1,2]{\fnm{Skylar} \sur{Sutherland}}\email{skylarsutherland@yale.edu}

\author[2,3]{\fnm{Bernhard} \sur{Egger}}\email{bernhard.egger@fau.de}

\author[2]{\fnm{Joshua} \sur{Tenenbaum}}\email{jbt@mit.edu}

\affil*[1]{\orgdiv{Department of Psychology}, \orgname{Yale University}, \orgaddress{\street{1 Hillhouse Ave}, \city{New Haven}, \postcode{06511}, \state{CT}, \country{USA}}}

\affil[2]{\orgdiv{Department of Brain and Cognitive Sciences}, \orgname{Massachusetts Institute of Technology}, \orgaddress{\street{77 Massachusetts Avenue}, \city{Cambridge}, \postcode{02139}, \state{MA}, \country{USA}}}

\affil[3]{\orgdiv{Department of Computer Science}, \orgname{Friedrich-Alexander-Universität Erlangen-Nürnberg}, \orgaddress{\street{Cauerstraße 11}, \city{Erlangen}, \postcode{91058}, \state{Bavaria}, \country{Germany}}}

\abstract{We propose a method for constructing generative models of 3D objects from a single 3D mesh and improving them through unsupervised low-shot learning from 2D images.  Our method produces a 3D morphable model that represents shape and albedo in terms of Gaussian processes. Whereas previous approaches have typically built 3D morphable models from multiple high-quality 3D scans through principal component analysis, we build 3D morphable models from a single scan or template.  As we demonstrate in the face domain, these models can be used to infer 3D reconstructions from 2D data (inverse graphics) or 3D data (registration).  Specifically, we show that our approach can be used to perform face recognition using only a single 3D template (one scan total, not one per person). We extend our model to a preliminary unsupervised learning framework that enables the learning of the distribution of 3D faces using one 3D template and a small number of 2D images. This approach could also provide a model for the origins of face perception in human infants, who appear to start with an innate face template and subsequently develop a flexible system for perceiving the 3D structure of any novel face from experience with only 2D images of a relatively small number of familiar faces.}

\maketitle

\section{Introduction}\label{sec:intro}

    3D generative models of objects are used in many computer vision and graphics applications.  Present methods for constructing such models typically require either significant amounts of 3D data processed through specialized pipelines, substantial manual annotation, or extremely large amounts of 2D data \cite{chaudhuri2020learning,egger20203d}.  We explore a novel approach that could provide a means to build generative models from very limited data: a single 3D object template (such as a single 3D scan of a face, or the average face in some population, or simply a hand-built coarse face mesh).  Our initial model is built using simple preprogrammed heuristics.  We then show that it can be improved using an unsupervised wake-sleep-like algorithm which learns statistical distributions of objects based on 2D observations, without relying on pretrained networks for feature point detection or face recognition.  The models we build are 3D morphable models (3DMMs) \cite{blanz1999morphable,egger20203d}, a type of generative model which creates samples by applying randomized shape and albedo deformations to a reference mesh. Traditionally, 3DMMs (e.g. \cite{paysan20093d, gerig2018morphable, li_learning_2017, booth_large_2018}) are built through principal component analysis (PCA) applied to datasets of 50 to 10,000 3D meshes produced by specialized (and expensive) 3D scanners \cite{egger20203d}.  Furthermore, a registration step is required to align the scans to a common topology.  In contrast, we use only a single scan or template, and so can eschew registration, an intrinsically ill-posed problem.
    
    Our approach uses the provided scan as our generative model's mean and smoothly deforms the scan as a surface in physical (3D) space and color (RGB) space using Gaussian processes.  Our shape deformation model follows that of \cite{luthi2017gaussian}.  We define the albedo deformations by combining analogous smooth albedo deformations on the mesh with smooth deformations on the surface defined by considering the mesh as a shape in RGB-space, with each vertex's location determined by its albedo rather than its position.  We initially define very generic Gaussian processes and add domain-specificity through correlation between color channels and bilateral symmetry.  Our models are fully compatible with PCA-based 3DMMs; the only difference is that our models' covariances are constructed through Gaussian processes rather than PCA.  As our 3DMMs use the same format and support the same operations as PCA-based 3DMMs, they can be used in existing pipelines to perform downstream tasks.  They can additionally be used to augment PCA-based 3DMMs \cite{luthi_chapter_2017}.

    This is, to the best of our knowledge, the most data-efficient procedure currently extant for constructing 3D generative models, and the sole procedure that only uses a single datapoint.  While the performance of our models is significantly poorer than that of PCA-based 3DMMs, they nevertheless perform surprisingly well given their data-efficiency.  While 3DMMs are a common prior in computer vision systems, their scalability is limited because their synthesis involves careful capture and modeling with category-specific domain knowledge.  Our method's data-efficiency obviates the need for large amounts of data capture, while our method's generality enables its use for any object class.  Finally, our approach minimizes the amount of sensitive personal data required to construct face 3DMMs.
    
    We also prototype an extension of our single-scan approach to a multi-scan setting by constructing mixture models of separate single-scan 3DMMs.  This can be seen as a generalization of kernel density estimation (KDE).  While performing inference with such a mixture model is more computationally expensive than with a PCA-based 3DMM, we demonstrate that the reconstruction quality obtained is much higher if the number of scans used in the models is low.  Furthermore, constructing this type of KDE-based model does not require correspondence between scans.
    
    In addition to extending our approach to multi-scan settings, we prototype a method for extending our single-scan 3DMMs on the basis of unsupervised low-shot learning from 2D images.  We do this by using our single-scan 3DMMs to perform analysis-by-synthesis \cite{yuille2006vision} on the 2D images, yielding a dataset of 3D reconstructions.  A new 3DMM can then be produced from this dataset through PCA.  Analysis-by-synthesis is performed using a three-stage pipeline: first, a CNN trained on synthetic data generated by the 3DMM is used to regress pose and lighting; second, a Markov chain Monte Carlo (MCMC) method is used to reconstruct the object within the single-scan 3DMM's eigenspaces; and third, a shape-from-shading strategy using the 3DMM as a source of regularization is used to reconstruct fine details. We demonstrate in the face domain that this approach can greatly improve the 3DMMs' visual quality using only a few hundred images.  Furthermore, we do not implicitly rely on supervision in the form of pretrained feature-point detectors or face recognition networks trained on labeled data; rather, we also bootstrap face alignment during learning.  Our approach is therefore unsupervised besides the single template and the heuristics that generate the initial model.
    
    We believe this approach to enhancing our 3DMM through the incorporation of unlabeled 2D data has applications not just for computer vision but also for the computational modeling of the development of face perception in infants.  The visual perception of infants is an area of key interest in cognitive science \cite{kellman2007infant} which is typically studied through psychophysical experiments, and rarely through computational models.  There are several theories explaining the development of face selective areas in the visual system and the preference very young infants have for faces \cite{slater1998newborn}.  One theory posits an innate subcortical face template, with the average face a likely candidate for such a template \cite{powell2018social}.  Studies of imitation in infants suggest some sort of basic face model might be present at birth \cite{meltzoff1989imitation}.  Psychophysical experiments with aftereffects in face perception also indicate the presence in the brain of a linear model of the space of faces that 3DMM's can quantitatively model \cite{leopold_prototype-referenced_2001, egger20203d}.  Furthermore, the adult brain's representation of face-space seems to be refined over the course of development \cite{valentine2016face}.  Recent neurological analysis of the face system in macaque monkeys supports analysis-by-synthesis using a 3DMM as a plausible underlying mechanism of face perception \cite{yildirim2020efficient}.  There is also strong evidence that human face perception improves drastically over the course of long-term development; despite the strong improvements in early vision, our face perception capabilities grow substantially in adulthood and peak in the 30s \cite{germine2011cognitive}.
    
    In the context of human cognitive development, our research also seeks to identify the minimal inductive biases that learning systems require to derive a 3DMM-based face perception framework from 2D images through unsupervised learning.  Our proposed unsupervised approach to learn a 3DMM is based on an inverse rendering framework along with a minimal 3D seed (which we demonstrate can be simpler than a full face scan) which we use to produce a weak generative model.  We argue that these mechanisms are plausibly innate in the human brain, with the average face representing a minimal innate template whose existence is indicated by infant experiments \cite{powell2018social}.  We then demonstrate that our framework can learn a rich 3DMM from 2D data.  This is the first fully unsupervised method for learning a 3DMM from 2D data, and the resulting model reaches broadly similar quality to existing 3DMMs learned from 2D data in a highly supervised manner.

    Although we demonstrate the applicability of our approach to other object categories, we focus our experiments on faces.  This is mainly because 3DMMs have historically been built for face modeling, so we can better compare our models to prior work in a face setting, and do so through well-established pipelines.  Our results with other object categories are harder to interpret, since our method is unique not only in its ability to generalize from a single datapoint, but also in its flexibility of object category.  However, our paper's methods may have their greatest relevance in domains outside of face perception since in the face domain high-quality 3DMMs built from 3D data are already widely extant.
    
    The main contributions of this work are the following:
    \begin{enumerate}
        \item We offer a novel albedo deformation model by combining surface-based and color-space-based kernels.
        \item We introduce a framework for 3DMM construction from a single 3D scan by extending an existing framework to build statistical shape models \cite{luthi2017gaussian} with our albedo deformation model.
        \item We evaluate our model on three downstream tasks, namely inverse rendering (2D to 3D registration), face recognition, and 3D to 3D registration, as well as the direct quality measures of specificity, generalization and compactness \cite{styner2003evaluation}.  We compare its performance with that of the 2019 (or, where relevant, 2017) Basel Face Model \cite{gerig2018morphable}, a state-of-the-art 3DMM produced from 200 3D scans.
        \item We build a prototype KDE-based face model from 10 face scans, and demonstrate that on a face recognition task it outperforms a PCA-based 3DMM built from the same 10 scans.
        \item We extend our framework on the basis of unsupervised low-shot learning from 2D images to enrich our simple model with image observations, demonstrating the feasibility of fully unsupervised learning of statistical 3DMMs.
    \end{enumerate}
    
\subsection{Related Work}\label{sec:related}

    The idea of building an axiomatic shape deformation model using Gaussian processes was previously explored in \cite{luthi2017gaussian}, which used such a deformation model as a prior for 3D registration tasks.  We extend this approach to include albedo along with shape by building Gaussian processes in RGB-space as well as physical space.  This enables its use as prior in an inverse graphics setting, and allows us to take albedo into account during registration.  \cite{kemelmacher20103d} presented a method for the 3D reconstruction of faces from 2D images through axiomatic deformation of a single 3D scan.  However, unlike our approach, this paper did not produce a generative model, and performed 3D reconstruction through shape-from-shading rather than probabilistic inference, using the 3D scan as purely as regularizer.  \cite{tegang2020gaussian} applied a Gaussian process intensity model in medical imaging for co-registration of CT and MRI images and for data augmentation. Other shape representation strategies (e.g. \cite{kilian2007geometric}) incorporate geodesic distances instead of Euclidean distances; while geodesic distances are beneficial in modeling motion and expression, since they are not easily transferable to color spaces we here focus on Euclidean distance.  \cite{ovsjanikov_exploration_2011} proposes a method for modeling variability in 3D datasets without correspondence by deforming a single template mesh.  However, unlike our work, \cite{ovsjanikov_exploration_2011} learns a nonlinear deformation model from a significant number of (unregistered) 3D scans through dimensionality reduction techniques, and so is inapplicable given only a single scan.  Furthermore, they only study 3D-to-3D reconstruction and it is unclear how their approach could be applied in a computer vision setting.
   
   While classically 3DMMs have been built from a collection of 3D scans, there are also several approaches that start from 2D data or combine 2D and 3D data.  Building a 3DMM solely from 2D data was first explored by \cite{cashman2012shape}.  Although they, like us, also start from a 3D mean shape as an initial template, their work neglects albedo.  Recently, methods to improve 3DMMs through 2D observations were proposed \cite{tewari2018self, tran2019learning}.  While they seek to build 3DMMs from 2D data, their approaches start with a full 3DMM built from 3D scans, and primarily refine the appearance model to increase flexibility.  Neither method offers a way to derive this initial model other than capturing 3D data and establishing correspondence between scans.  \cite{tran2019towards} further extended these ideas to incorporate nonlinear models so as to overcome the limitations inherent in the linearity of classical 3DMMs.  In contrast to these works aiming to build a 3DMM from a large collection of 2D data and an initial 3DMM, our work focuses on building a 3DMM from just a single 3D scan.  Such a model could be used as an initial model for the 2D learning strategies discussed above.
   
   Additionally, some recent work has focused on the problem of the unsupervised learning of 3D generative models from a large 2D training corpus \cite{szabo2019unsupervised, wu2020unsupervised} or from depth data \cite{abrevaya_multilinear_2018}.  The 3D generative models learned by these approaches do not disentangle illumination and albedo (or neglect albedo entirely, as in \cite{abrevaya_multilinear_2018}), and do not preserve correspondence, making them difficult to interpret.  Furthermore, this means that they are incompatible with existing 3DMM-based pipelines; in contrast, generative models produced through our approach can be used interchangeably with PCA-based 3DMMs.
   
   \cite{tewari2020learning} was the first to propose a complete 3DMM learned from 2D images and video data through self-supervised learning, using an average 3D face for initialization.  This paper is more directly comparable to our work.  However, we show that the average face is already sufficient to produce a usable 3DMM, without any 2D data.  While we do prototype extensions of our 3DMMs on the basis of 2D data, we do so using far less data than \cite{tewari2020learning}: we use only static images, use many orders of magnitude fewer images, and do not rely on pretrained face detectors, feature point detectors or face recognition networks. Their model however also incorporates facial expressions through video supervision which are omitted in our proof of concept.
   
   Other works have focused on extending 3D morphable models beyond a linear latent space \cite{ranjan_generating_2018, bouritsas_neural_2019, tran2019towards}.  In contrast, we use a traditional linear latent space and rather focus on how such latent spaces can be learned.  For additional work on applications of 3DMMs and shape and albedo representations used with 3DMMs, we refer to \cite{egger20203d}.
   
   Our analysis-by-synthesis method presented in Section~\ref{sec:learning} is closely related to a number of prior works.  The first stage of our method, a CNN trained on 3DMM-generated synthetic data for pose and lighting regression, is similar to the Efficient Inverse Graphics network of \cite{yildirim2020efficient} and previous work on regressing 3DMM parameters directly from images \cite{tuan2017regressing}.  The second stage, an MCMC method for shape and albedo regression within a 3DMM's shape and albedo subspaces, is similar to the MCMC method presented in \cite{schonborn2017markov} (and we use \cite{schonborn2017markov}'s method directly in other parts of the paper).  The final stage, a 3DMM-regularized shape-from-shading strategy, is loosely similar to the approaches of \cite{kemelmacher-shlizerman_3d_2011} and \cite{patel_driving_2012}; however, the specific combination of these approaches, and their use case, is original to our paper.
   
   In addition to 3D morphable models, our work can also be connected with \textit{shape-from-template} approaches to 3D vision.  These approaches typically address the following problem: given a reference mesh, an input image, and a set of dense (pixel-level) correspondences between the input image and a rendering of the mesh \cite{bartoli_template_based_2012, ostlund_laplacian_2012, brunet_monocular_2011, malti_pixel_based_2011, moreno_noguer_exploring_2010, salzmann_linear_2011} or with the mesh directly \cite{moreno_noguer_capturing_2009, salzmann_closed_form_2008}, deform the mesh to match the input image.  Restrictions on the allowed deformations (e.g. isometry or conformality) make this problem well-posed and sometimes solvable analytically.  This framing means that shape-from-template approaches are rarely applicable without dense 2D correspondence annotations and generally ignore albedo.  Shape-from-template approaches that do not require dense 2D correspondence have typically previously relied on additional 3D or video data and still do not fully model albedo \cite{yu_direct_2015, salzmann_local_2008, shaji_simultaneous_2010}.  In contrast, our approach can infer 3D reconstructions from single images reliably using only a small set of landmarks (sparse 2D correspondence) for localization, and our unsupervised learning approach is capable of fully unsupervised (albeit much less reliable) 3D face reconstruction.  We furthermore separate albedo and illumination and fully incorporate albedo in our deformations.  Furthermore, shape-from-template approaches have no way to incorporate statistical information about the distribution of 3D objects likely to be observed, whereas we demonstrate our approach can incorporate statistical learning.
   
   This work extends our previous conference paper to incorporate unsupervised learning \cite{sutherland2021building}.

\section{Methods}\label{sec:methods}

    A 3DMM consists of a shape model and an albedo model; samples from a 3DMM are meshes with a common topology, with the position and albedo of each vertex generated by the 3DMM's shape and albedo models, respectively \cite{egger20203d}.  Our framework represents samples from the shape and albedo models as deformations of a vertex-colored mesh that defines both the topology of all samples and the mean of the shape and albedo distributions.  Our approach uses the 3D scan as the mean of the resulting 3DMM.  We define the shape and albedo models in terms of Gaussian processes, each consisting of a mean and a covariance kernel \cite{luthi2017gaussian, rasmussen2003gaussian}.  While this method's performance depends on the choice of mean mesh, PCA-based 3DMMs face the same issue since registration likewise requires a choice of common topology.

    We define a Gaussian process $g$ as a pair $(\mu, \Sigma)$, where $\mu$ is the mean of the Gaussian process and $\Sigma$ is the covariance kernel of the Gaussian process; $\mu$ is a function from $A$ to $\R^n$ for some set $A$ and constant $n$, and $\Sigma$ is a positive-definite function from $A^2$ to $\R^{n \times n}$, where $\R^{n \times n}$ is the space of $n$-by-$n$ matrices.  In our case, for both the shape and albedo models, $A$ is the set of mesh vertices, and $n = 3$.  A sample from the shape model maps $A$ to positions in $\R^3$, whereas a sample from the albedo model maps $A$ to RGB values, represented as vectors in $\R^3$.  We represent our shape and albedo kernels using Mercer decomposition computed through the Nystr\"om method \cite{rasmussen2003gaussian, luthi2017gaussian}.

\subsection{Shape Covariance Kernels}\label{subsec:shapekernels}

    We follow the approach of \cite{luthi2017gaussian}: defining covariance kernels which give a high correlation between nearby points and a low correlation between distant points.  The most straightforward way to do this is with physical distance.  Our shape kernels are based on radial basis function kernels  \cite{rasmussen2003gaussian, luthi2017gaussian}.

    A function $f: A^2 \to \R$ is positive-definite if the matrix $M$ defined by $M_{i, j} = f(x_i, x_j)$ is positive-semidefinite for any $x_1, \ldots, x_n \in \R$ \cite{mercer_functions_1909}.  This definition can be extended to matrix-valued kernels by letting $M_{i, j}$ represent a block submatrix of $M$ instead of an entry of $M$ \cite{rasmussen2003gaussian, luthi2017gaussian}.  Since the set of positive-semidefinite matrices is closed under addition and positive scalar multiplication \cite{horn_matrix_2012}, so are matrix-valued kernels.  In order to create a kernel with a coarse-to-fine structure, possessing strong short-range correlations and weaker long-range correlations, we define our shape kernel as a linear combination of radial basis function kernels.  Letting $\Sigma_{s, \sigma}$ represent the radial basis function kernel defined using physical distance as its metric and scale $\sigma$, we define the family of scalar kernels $\Sigma_{\texttt{std}}(a, b, c, A, B, C) = a \Sigma_{s, A} + b \Sigma_{s, B} + c \Sigma_{s, C}$.  We here let $\Sigma_0 = \Sigma_{\texttt{std}}(a_s, b_s, c_s, A_s, B_s, C_s)$, where $a_s$, $b_s$, $c_s$, $A_s$, $B_s$, and $C_s$ are hyperparameters (listed in Section~\ref{subsec:hyperparameters}).
    
    In order to represent 3D deformations, we must multiply scalar kernels by $3$-by-$3$ matrices.  Since we wish for deformations in $x$, $y$, and $z$ to be uncorrelated, we simply multiply by $I_3$, the $3$-by-$3$ identity matrix.  Thus, our standard shape kernel is $K_s = I_3 \Sigma_0$.  One limitation of this kernel is that it does not encode bilateral symmetry.  Many object categories, including faces, are bilaterally symmetric.  In order to add symmetry to this kernel, we wish to make the deformations applied to points on opposite sides of the object closely correlated in the up-down and forward-back axes and strongly anticorrelated in the left-right axis \cite{morel2016generative}.  To define such kernels, let $\Phi_m \in \R^{3 \times 3}$ be the matrix which, considered as a linear transformation applied to points in physical space, negates a point's left-right component (where left and right are defined relative to the scan).  Then our symmetric shape kernel is defined as $K_s^{\texttt{sym}} = I_3 \Sigma_0(x, y) + \alpha \Phi_m \Sigma_0(x, \Phi_m(y))$, where $\Phi_m(y)$ denotes applying $\Phi_m$ as a linear transformation to $y$'s position in $\R^3$, and $\alpha$ is a hyperparameter (listed in Section~\ref{subsec:hyperparameters}) \cite{morel2016generative}.

\subsection{Albedo Covariance Kernels}\label{subsec:albedokernels}

    What we principally desire in an albedo kernel is that deformations applied to different areas should be highly correlated if and only if the areas are related.  Unlike shape deformations, albedo deformations in general need not be spatially continuous, and so a global notion of similarity is needed in addition to physical proximity.  We measure the similarity of mesh vertices by combining their distance in physical space with their distance in albedo space.
    
    Physical distance is a straightforward way of assessing similarity.  We define a physical distance-based albedo kernel similarly to $K_s$.  Specifically, we define $K_{a, \texttt{xyz}} = I_3 \Sigma_{\texttt{xyz}}$, where $\Sigma_{\texttt{xyz}} = \Sigma_{\texttt{std}}(a_a, b_a, c_a, A_a, B_a, C_a)$, with hyperparameters $a_a$, $b_a$, $c_a$, $A_a$, $B_a$, and $C_a$ listed in Section~\ref{subsec:hyperparameters}.  Samples from $K_{a, \texttt{xyz}}$ represent deformations in RGB-space, not position.  However, this kernel neglects some kinds of similarity.  For instance, in a human face, a point on a lip is more similar to another point on a lip than it is to an equidistant point on a cheek; more generally, many objects exhibit part-based similarity in addition to distance-based similarity.  Color-space distance provides us with an estimate of part-based similarity that does not depend on explicit part annotations.  Just as the distances between mesh points in physical space (in the mean) constitute a metric on the set of mesh points, so do the Euclidean distances between mesh points' albedos, represented as RGB values and considered as points in $\R^3$.  Using this alternate metric, we may define another family of radial basis function kernels, which we term $\Sigma_{a, \sigma}$ for $\sigma \in \R$. We then define the alternate albedo kernel $K_{a, \texttt{rgb}} = I_3 \Sigma_{\texttt{rgb}}$, where $\Sigma_{\texttt{rgb}} = d \Sigma_{a, D}$, with hyperparameters $d$ and $D$ (listed in Section~\ref{subsec:hyperparameters}).
    
    To use both local and global information, we average these kernels.  A core contribution here is the combined kernel $K_a = 0.5 (K_{a, \texttt{xyz}} + K_{a, \texttt{rgb}})$.  This kernel takes into account both the differences in position and differences in albedos between points on the mesh, and can thus relatively robustly assess whether different parts of the object are parts of the same component.

    As stated, all three of our albedo kernels are products of a scalar-valued kernel with $I_3$. Multiplying by a different matrix enables us to incorporate domain knowledge about an object category's common albedos by correlating the different color channels (red, green, and blue).  In particular, as a very rough approximation to human skin tones, we introduce additional kernels $K_{a, \texttt{xyz}}^{\texttt{cor}}$ and $K_{a, \texttt{rgb}}^{\texttt{sym}}$, depending, respectively, on physical and RGB-space distance.  Letting
    \begin{equation}
        M_x = \begin{bmatrix} 1 & x & x \\ x & 1 & x \\ x & x & 1 \end{bmatrix}
    \end{equation}
    we define $K_{a, \texttt{xyz}}^{\texttt{cor}} = M_{\beta} \Sigma_{\texttt{xyz}}$ and $K_{a, \texttt{rgb}}^{\texttt{sym}} = M_{\gamma} \Sigma_{\texttt{rgb}}$, where $\beta$ and $\gamma$ are hyperparameters (listed in Section~\ref{subsec:hyperparameters}).
    
    To add further domain knowledge we create additional albedo kernels that incorporate bilateral symmetry.  Since the albedo of a member of a bilaterally symmetric object class is essentially bilaterally symmetric, $K_{a, \texttt{rgb}}^{\texttt{sym}}$ is already symmetric in practice.  However, the physical-distance-based albedo kernels can be symmetrized via a process analogous to that used for the shape kernels, with the difference that we do not wish to negate left-right deformations located on opposite sides of the object.  We choose to consider color channel correlations and symmetry simultaneously, and so define $K_{a, \texttt{xyz}}^{\texttt{sym}}(x, y) = K_{a, \texttt{xyz}}^{\texttt{cor}}(x, y) + \alpha K_{a, \texttt{xyz}}^{\texttt{cor}}(x, \Phi_m y)$, and define $K_a^{\texttt{sym}} = 0.5 (K_{a, \texttt{rgb}}^{\texttt{sym}} + K_{a, \texttt{xyz}}^{\texttt{sym}})$.
    
    To attempt to separate the roles played by symmetry and color-channel correlation, in the supplementary material we also present results with albedo kernels that have correlated color channels but lack symmetry.

\subsection{Kernel Density Estimation}\label{subsec:kde}

    One limitation of our approach is that it provides no way to leverage the information present in multiple scans.  However, an extension of our approach can be used in a setting where multiple scans are available.  To construct a model from multiple scans, we create single-scan 3DMMs for each scan separately, and then build a mixture model from the different 3DMMs built from each individual scan, resulting in a non-parametric 3DMM-based model.  This essentially amounts to an extension of kernel density estimation (KDE), where Gaussian processes replace uniform Gaussian distributions in the definitions of each mixture component, providing a non-uniform noise model.
    
    An advantage of this kernel density estimation approach is that, unlike PCA, it does not require dense correspondence between scans.  This could enable the creation of 3DMM-based generative models of object categories where many 3D scans exist but where establishing dense correspondence is impossible (e.g. chairs).  However, the non-parametric nature of a KDE-based model means that, unlike a PCA-based 3DMM, the amount of computation required to perform inference grows with the number of scans.

\subsection{Learning from 2D Data}\label{sec:learning}

    The single-scan 3DMMs constructed above, while usable in some downstream tasks, remain very far from object categories' true distributions.  In the face domain, we further demonstrate how our 3DMMs can be augmented through unsupervised low-shot learning from 2D data.  This is done by first producing 3D reconstructions of faces from 2D images through inverse graphics using the initial single-scan 3DMM, and then applying PCA to the resulting dataset.  This cannot be done using the analysis-by-synthesis method of Sch\"onborn et al. \cite{schonborn2017markov} (which we use in Section~\ref{sec:experiments}) for two reasons.  Firstly, Sch\"onborn et al.'s method relies on manual landmark annotations.  Secondly, it can only produce reconstructions which lie within the support of the single-scan 3DMM, and applying PCA to a dataset of such reconstructions will yield a 3DMM whose support is equal to or a subset of that of the single-scan 3DMM.  For these reasons we produce reconstructions using a new unsupervised analysis-by-synthesis method, which is outlined below.
    
    \begin{figure}[H]
        \begin{center}
            \includegraphics[width=0.99\linewidth]{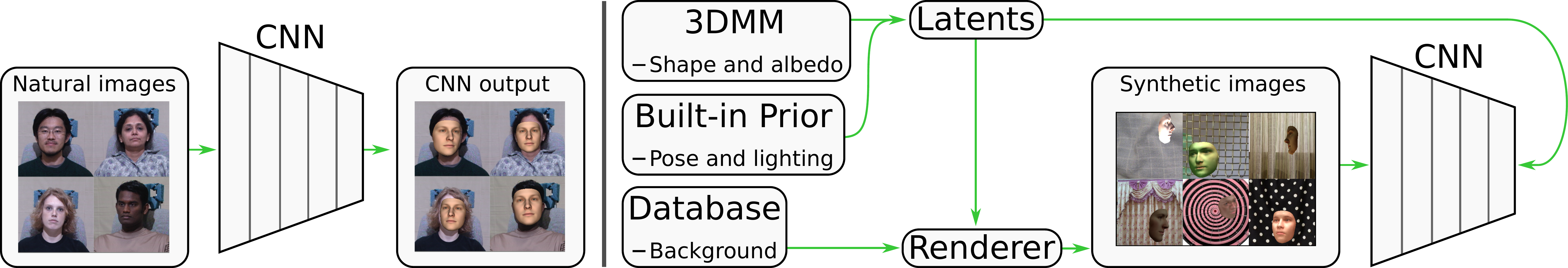}
        \end{center}
        \caption{The first stage of our inference pipeline.  A convolutional neural network is trained on synthetic data rendered from our single scan model to regress the location and pose of faces in the input images.}
        \label{fig:learning_cnn}
    \end{figure}
    
    \begin{figure}[H]
        \begin{center}
            \includegraphics[width=0.99\linewidth]{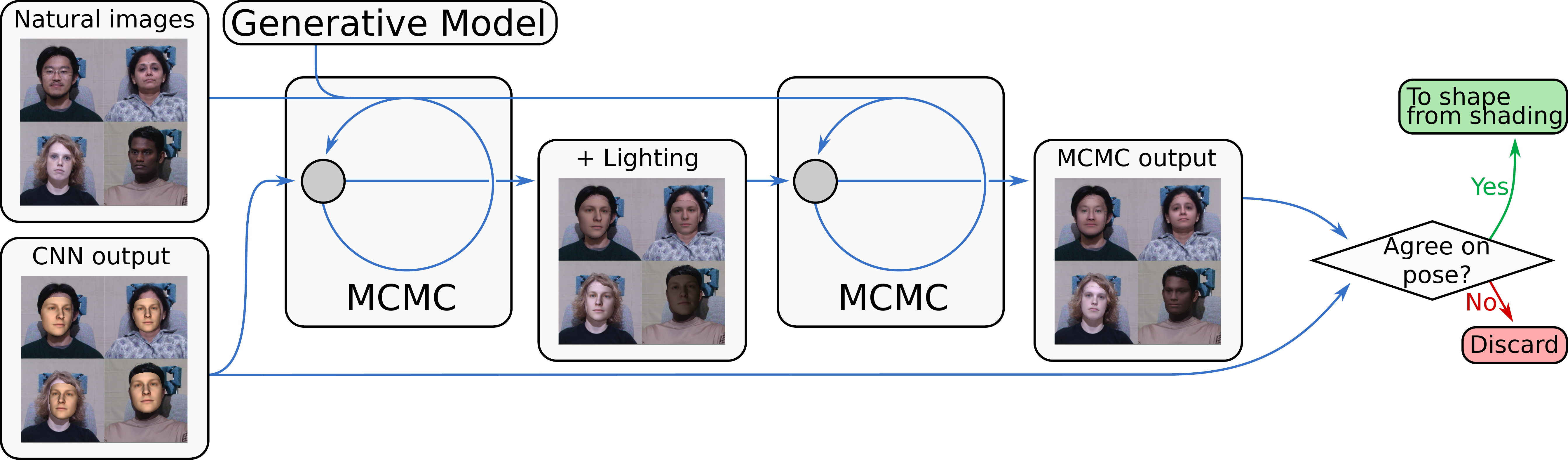}
        \end{center}
        \caption{The second stage of our inference pipeline.  A Markov chain Monte Carlo (MCMC) process is used to infer the shape and albedo of the face along with environmental illumination, as well as refine the regressed pose.  If the MCMC process does not approximately preserve the 2D silhouette of the rendered face the face reconstruction is discarded.}
        \label{fig:learning_mcmc}
    \end{figure}

    \begin{figure}[H]
        \begin{center}
            \includegraphics[width=0.7\linewidth]{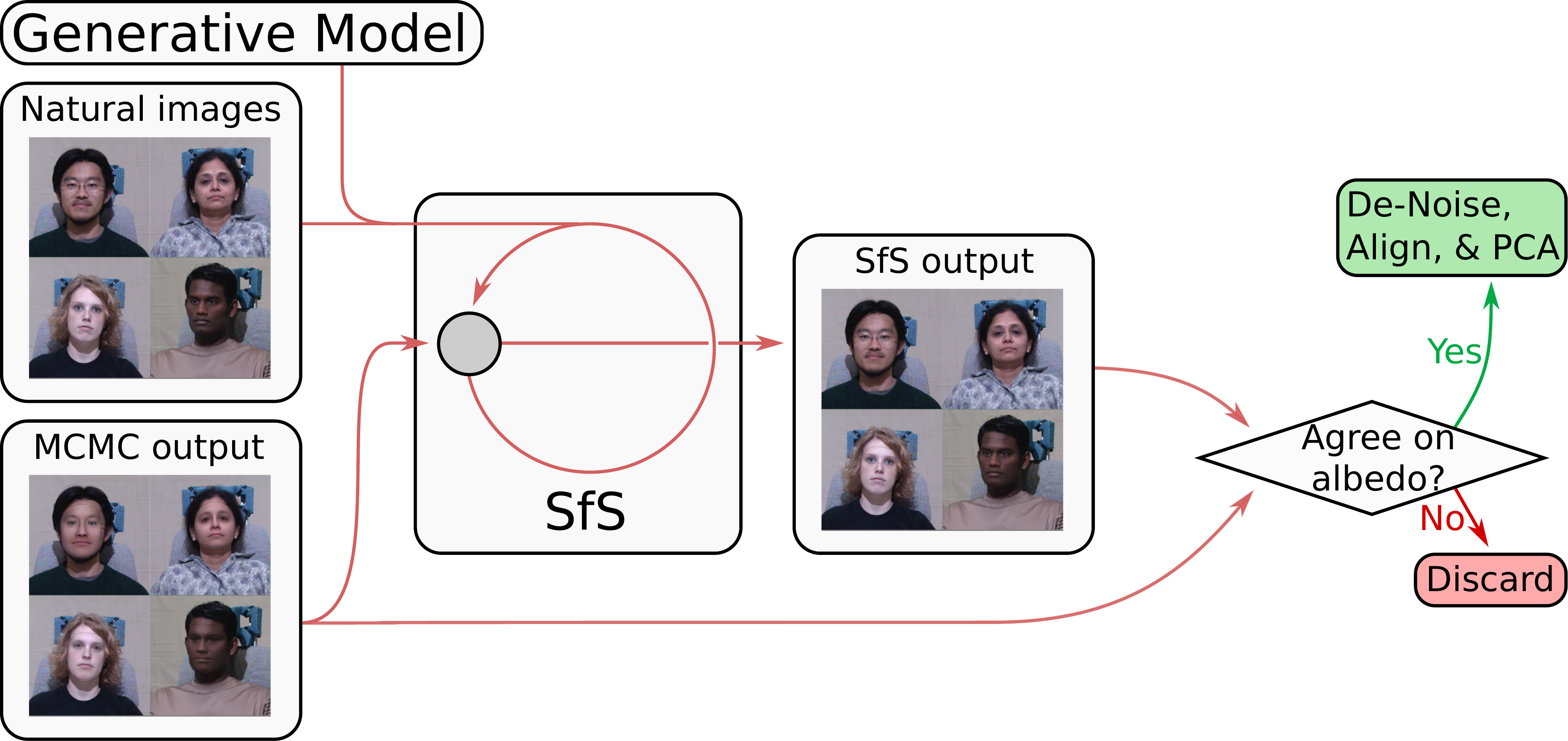}
        \end{center}
        \caption{The third stage of our inference pipeline.  A shape-from-shading strategy is used to infer the fine details of the shape and albedo of the face along with refined environmental illumination.  Once the shape-from-shading process is complete, denoising and alignment steps are applied as post-processing.  If the shape-from-shading process does not approximately preserve albedo, the reconstruction is discarded,}
        \label{fig:learning_sfs}
    \end{figure}

    To produce 3D face reconstructions from 2D images, we use a three-stage inference process.  First, as illustrated in Figure~\ref{fig:learning_cnn}, a convolutional neural network (CNN) trained on synthetic data regresses the face position and orientation and scene lighting, similarly to \cite{yildirim2020efficient} but with non-face scene parameters regressed rather than a 3DMM's principal components.  We use a pretrained ResNet50-v2 network \cite{he2016identity}. The network is trained with synthetic data that incorporates variation in both shape and albedo as well as camera and illumination parameters. The aim of this first step is to roughly align the face and estimate the illumination.
    
    Next, as illustrated in Figure~\ref{fig:learning_mcmc}, these parameters are used to initialize a MCMC process broadly similar to that of \cite{schonborn2017markov} (without landmarks and with a somewhat different proposal distribution) that produces a 3D reconstruction of the face's shape and albedo within the 3DMM's subspaces while also inferring lighting and refining the estimated pose.  The precise hyperparameters used in the proposal generation distribution are slightly altered due to the use of a different underlying computational framework, and unlike in the case of \cite{schonborn2017markov}, we incorporate a canonical prior on pose and lighting parameters (which is also used to generate synthetic training data).  However, the basic structure of \cite{schonborn2017markov}'s proposal distribution---a coarse-to-fine mixture model of Gaussian drift hypotheses---is preserved.  We use initial $n_1$ initial MCMC steps to estimate the lighting parameters only, followed by $n_2$ MCMC steps to estimate the other parameters (along with refined lighting parameters).  $n_1$ and $n_2$ are hyperparameters listed in Section~\ref{subsec:hyperparameters}.
    
    Finally, as illustrated in Figure~\ref{fig:learning_sfs}, a shape-from-shading strategy is used to reconstruct the face's fine details outside of the 3DMM's shape and albedo subspaces.  Since shape-from-shading is an ill-posed problem, shape-from-shading approaches inherently require some source of regularization \cite{zhang_shape--shading_1999}.  We use our 3DMM as a source of regularization, penalizing reconstructions both based on their distance from the 3DMM's shape and albedo subspaces, and the probability the 3DMM assigns the reconstructions' projections into those subspaces.  Optimization is then performed using gradient descent, using $n_3$ sequential gradient descent steps, where $n_3$ is a hyperparameter listed in Section~\ref{subsec:hyperparameters}.  This makes the shape-from-shading process similar to maximum \textit{a posteriori} optimization using differentiable rendering, and is loosely similar to the approaches of \cite{kemelmacher-shlizerman_3d_2011} and \cite{patel_driving_2012}.
    
    Once a dataset of detailed 3D reconstructions has been produced, a new 3DMM is constructed by applying PCA to this dataset, and the learning process repeats in the next iteration.  Before PCA is applied, a denoising step is applied to the shape of each mesh to remove any spikes introduced by the shape-from-shading process, and an alignment step using the algorithm of \cite{umeyama_least-squares_1991} is applied.  Once a new 3DMM has been produced, the CNN used to initialize pose and lighting estimation is finetuned through retraining on new synthetic data.  The new synthetic dataset used to train the CNN does not consist solely of samples from the newly constructed 3DMM, but rather is a mixture of samples from all the 3DMMs produced throughout the learning process.  Improvements in the 3DMM also improve the accuracy of the MCMC and shape-from-shading steps, since these are both model-in-the-loop processes.
    
    Quality control at each of the steps in our framework is essential for the learning of a 3DMM. A single bad reconstruction can introduce significant errors and severe artifacts into the learned model. We therefore implemented simple quality control heuristics that ensure that only the very best reconstructions end up in the resulting model. Misalignments of the face in the image by the CNN can yield highly inaccurate MCMC reconstructions, and similar misalignments during the MCMC process can yield gross errors in the shape-from-shading reconstruction.  For this reason, we discard poor fits during the process using simple heuristics.  Specifically, if the MCMC process does not approximately preserve the 2D silhouette of the face reconstruction, or if the shape-from-shading strategy does not approximately preserve the albedo of the face reconstruction, the reconstruction is discarded as a probable failure.  At the end of the MCMC process, the 2D silhouettes of the rendered fits are compared and if the ratio of pixels in both silhouettes to pixels in either silhouette is not at least $r_1$ (a hyperparameter), the fit is discarded.  Similarly, at the end of the shape-from-shading process, we compute the average distance between the albedo of each vertex in the mesh before and after the shape-from-shading process, and if this exceeds a threshold $r_2 - n r_3$ the fit is discarded, where $r_2$ and $r_3$ are hyperparameters and $n$ is the number of previously performed wake-sleep iterations.  While the precise number of reconstructions that pass these quality control steps varies, it is generally very low, and the quality control steps can be viewed as selecting the very best reconstructions.
    
\subsection{Choices of Hyperparameters}\label{subsec:hyperparameters}

    We choose as hyperparameters $a_s = 7$, $b_s = 5$, $c_s = 3$, $A_s = 100$, $B_s = 50$, $C_s = 10$, $a_a = 0.02$, $b_a = 0.01$, $c_a = 0.01$, $A_a = 500$, $B_a = 20$, $C_a = 2$, $d = 0.015$, $D = 0.15$, $\alpha = 0.7$, $\beta = 0.9375$, $\gamma = 0.95$, $n_1 = 1000$, $n_2 = 10000$, $n_3 = 5000$, $r_1 = 0.625$, $r_2 = 8$, and $r_3 = 0.5$.  Importantly, these parameters have generally not been extensively tuned, which is reflected in the fact that we use the same kernels for faces, birds and fish.  Our core idea is simply to combine radial basis function kernels at three different scales and magnitudes so as to incorporate global as well as local flexibility.  Hyperparameters representing physical distances (namely $a_s$, $b_s$, $c_s$, $A_s$, $B_s$, $C_s$, $A_a$, $B_a$, and $C_a$) have units of millimeters.  We represent RGB values as points in $[0, 1]^3$, and $a_a$, $b_a$, $c_a$, $d$, $D$, $r_2$ and $r_3$ represent distances or magnitudes in color space using this unit system.

\section{Experiments}\label{sec:experiments}

    We produce a set of 3DMMs from our kernels using the average face of the 2019 Basel Face Model \cite{gerig2018morphable} as our reference mesh.  These are listed with their corresponding kernels in Table~\ref{tab:kernels}.  These 3DMMs have the same mean as the 2019 Basel Face Model, and so comparing their performance with that of the 2019 Basel Face Model constitutes a direct comparison of our axiomatic Gaussian process-based covariance kernels with the learned covariance model of the 2019 Basel Face Model.  We also produce 3DMMs by combining our kernels with face scans provided with the 2009 Basel Face Model \cite{paysan20093d}.  We assess these models' performance on downstream tasks where 3DMMs are often used, namely inverse graphics (in Section~\ref{subsec:ir}) and registration (in Section~\ref{subsec:registration}), and directly compare these 3DMM's specificity and generalization on real faces with that of the 2017 Basel Face Model \cite{gerig2018morphable} in Section~\ref{sec:specgen}.  In Section~\ref{sec:learnedmodels} we show samples and face reconstructions from 3DMMs learned from 2D data using our wake-sleep approach and compare them with the LeMoMo model of Tewari et al. \cite{tewari2020learning}.  In Section~\ref{subsec:other_objects} we experiment with simple 3DMMs of birds and fish.

    \begin{table}[ht]
        \centering
        \begin{tabular}{|l|c|c|}
            \hline
            name & shape kernel & albedo kernel\\
            \hline
            standard-full & $K_s$ & $K_a$\\
            \hline
            standard-RGB & $K_s$ & $K_{a, \texttt{rgb}}$\\
            \hline
            standard-XYZ & $K_s$ & $K_{a, \texttt{xyz}}$\\
            \hline
            symmetric-full & $K_s^{\texttt{sym}}$ & $K_a^{\texttt{sym}}$\\
            \hline
            symmetric-RGB & $K_s^{\texttt{sym}}$ & $K_{a, \texttt{rgb}}^{\texttt{sym}}$\\
            \hline
            symmetric-XYZ & $K_s^{\texttt{sym}}$ & $K_{a, \texttt{xyz}}^{\texttt{sym}}$\\
            \hline
        \end{tabular}
        \caption{Our Gaussian processes for modeling faces.}
        \label{tab:kernels}
    \end{table}
    
\begin{figure}[H]
        \begin{center}
            \begin{tabular}{|c|c|c|c|c|c|c|c|c|c|}
                \hline
                $K_s$ & $K_s^{\texttt{sym}}$ & $K_{a, \texttt{xyz}}$ & $K_{a, \texttt{rgb}}$ & $K_a$ & $K_{a, \texttt{xyz}}^{\texttt{sym}}$ & $K_{a, \texttt{rgb}}^{\texttt{sym}}$ & $K_a^{\texttt{sym}}$\\
                \hline
                \includegraphics[width=0.07\linewidth]{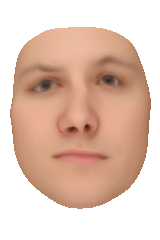} &
                \includegraphics[width=0.07\linewidth]{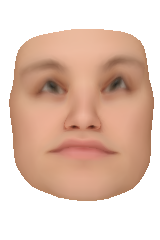} &
                \includegraphics[width=0.07\linewidth]{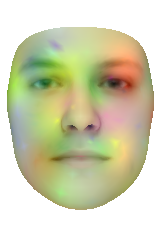} &
                \includegraphics[width=0.07\linewidth]{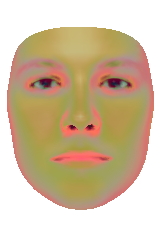} &
                \includegraphics[width=0.07\linewidth]{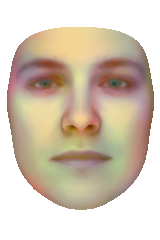} &
                \includegraphics[width=0.07\linewidth]{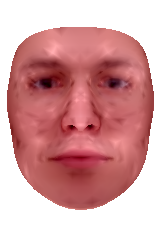} &
                \includegraphics[width=0.07\linewidth]{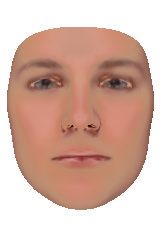} &
                \includegraphics[width=0.07\linewidth]{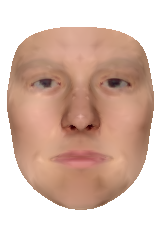}\\
                \hline
                \includegraphics[width=0.07\linewidth]{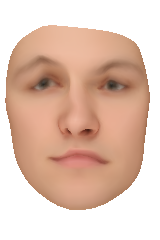} &
                \includegraphics[width=0.07\linewidth]{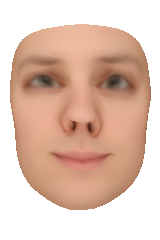} &
                \includegraphics[width=0.07\linewidth]{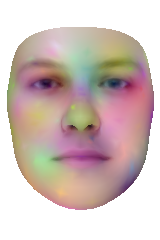} &
                \includegraphics[width=0.07\linewidth]{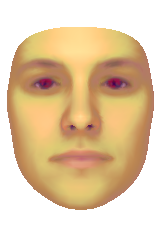} &
                \includegraphics[width=0.07\linewidth]{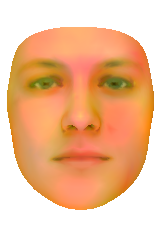} &
                \includegraphics[width=0.07\linewidth]{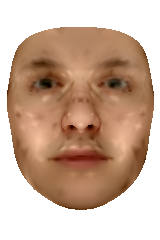} &
                \includegraphics[width=0.07\linewidth]{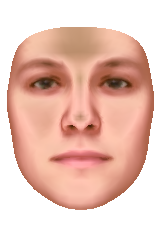} &
                \includegraphics[width=0.07\linewidth]{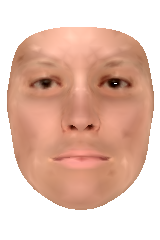}\\
                \hline
                \includegraphics[width=0.07\linewidth]{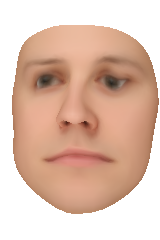} &
                \includegraphics[width=0.07\linewidth]{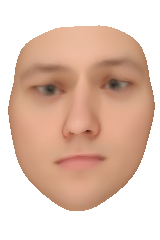} &
                \includegraphics[width=0.07\linewidth]{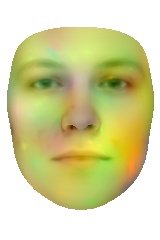} &
                \includegraphics[width=0.07\linewidth]{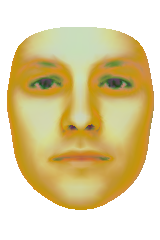} &
                \includegraphics[width=0.07\linewidth]{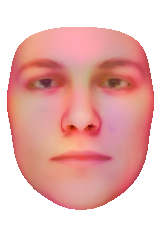} &
                \includegraphics[width=0.07\linewidth]{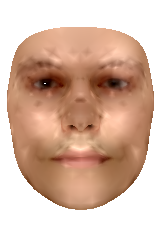} &
                \includegraphics[width=0.07\linewidth]{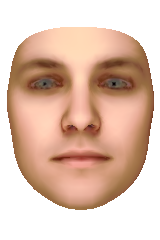} &
                \includegraphics[width=0.07\linewidth]{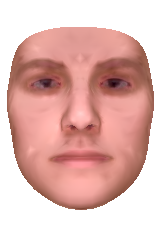}\\
                \hline
            \end{tabular}
        \end{center}
        \caption{Three random samples from each of the shape and albedo kernels applied to the mean of the 2019 Basel Face Model \cite{gerig2018morphable} and rendered under ambient illumination.  The first two columns are the two shape kernels, while the remaining eight columns are the albedo kernels.}
        \label{fig:kernel_illustrations}
    \end{figure}

    In Figure~\ref{fig:kernel_illustrations}, we show samples from our various shape and albedo kernels, applied to the mean of the 2019 Basel Face Model \cite{gerig2018morphable}.  While these samples are clearly non-naturalistic, this does not invalidate the results of Section 3 of our main paper.  \textbf{We make no claim that our initial 3DMMs accurately model the distribution of human faces; rather, we claim that they are of sufficient quality to be useful in a machine vision context and that we can improve their quality based on a few 2D observations.}
    
\subsection{Inverse Rendering}\label{subsec:ir}

    One of the most direct ways to assess the value of our model is to apply it in an analysis-by-synthesis setting \cite{yuille2006vision}.  Using our 3DMMs as priors on 3D meshes, we can perform inverse rendering to reconstruct 3D meshes from 2D images through approximate posterior inference \cite{schonborn2017markov}.  We use a spherical harmonics lighting model, as in \cite{zivanov_human_2013}, and a pinhole camera model, as in \cite{blanz1999morphable}.  Since no synthetic image will ever exactly match a natural image, we treat foreground pixels as subject to Gaussian noise and background pixels as sampled from the input image, following the method of \cite{schonborn2015background}.
    
    To perform inference, we use the MCMC method presented in \cite{schonborn2017markov}.  Specifically, we use Gaussian drift proposals to update pose, perform closed-form estimation of illumination, and use Gaussian drift proposals applied in the 3DMM's low-dimensional eigenspaces to update the mesh itself.  In order to locate the face in the image we constrain pose using landmark annotations provided with each image.  Although we generated these landmark annotations manually, they could also have been obtained automatically using existing tools (e.g. OpenPose \cite{cao_openpose_2019}).

    \begin{figure}[H]
        \begin{center}
            \includegraphics[width=0.95\linewidth]{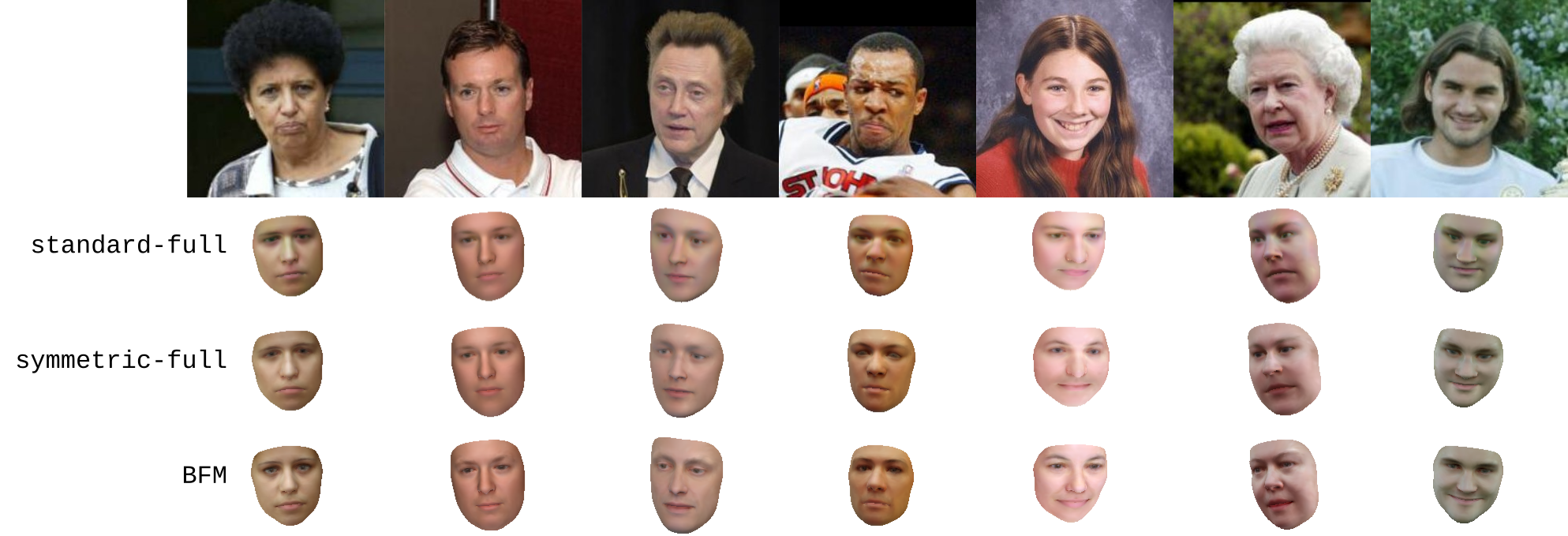}
        \end{center}
        \caption{Reconstructions produced from natural images using various 3DMMs.  The first row shows the natural images used as input, while the remaining rows show the reconstructions inferred using different 3DMMs.  The standard-full and symmetric-full models were produced using the mean of the 2019 Basel Face Model \cite{gerig2018morphable} as template.}
        \label{fig:lfw_results}
    \end{figure}

    One analysis-by-synthesis task is to reconstruct 3D face meshes from natural images, render the results and compare them with said natural images.  We here perform this task on images from the Labeled Faces in the Wild dataset \cite{huang2008labeled} and show in Figure~\ref{fig:lfw_results} the reconstructions produced using the standard-full and symmetric-full 3DMMs (as defined in Table~\ref{tab:kernels}), as well as the reconstructions that our inverse graphics pipeline produces using the 2019 Basel Face Model \cite{gerig2018morphable}.  As Figure~\ref{fig:lfw_results} demonstrates, all of these 3DMMs produce high-quality reconstructions.
    
    In addition to the models we define using the mean of the 2019 Basel Face Model, we construct additional 3DMMs using the symmetric kernel and ten scans provided with the 2009 Basel Face Model \cite{paysan20093d} as different means.  We name these models symmetric-$x$, where $x$ is the ID number of the scan (\texttt{001}, \texttt{002}, \texttt{006}, \texttt{014}, \texttt{017}, \texttt{022}, \texttt{052}, \texttt{053}, \texttt{293}, or \texttt{323}).  Reconstructions produced by these 3DMMs can be found in the supplementary material, along with side views of our reconstructions.  To assess our 3DMMs' performance in an inverse graphics setting where the choice of prior gains importance, the supplementary material also includes reconstructions of partially occluded faces produced with the occlusion-aware MCMC method described in \cite{egger2018occlusion}.  All models again yield similar reconstruction quality.

    \begin{table}[ht]
        \centering
        \begin{tabular}{|llll|} 
            \hline
            angle                               & $15^\circ$        & $30^\circ$        &  $45^\circ$ \\
            probe id                            & \texttt{140\_16}  & \texttt{130\_16}  & \texttt{080\_16} \\ 
            \hline
            standard-full                       & 84.7              & 69.9              & 54.2 \\
            standard-RGB                        & 76.3              & 57.8              & 28.9 \\
            standard-XYZ                        & 77.1              & 62.7              & 35.7 \\
            symmetric-full                      & \textbf{93.2}     & \textbf{85.9}     & \textbf{72.3} \\
            symmetric-RGB                       & 73.5              & 61.4              & 40.2 \\
            symmetric-XYZ                       & 73.1              & 58.6              & 44.2 \\
            \hline
            symmetric-\texttt{001}              & 78.7              & 62.2              & 49.8 \\
            symmetric-\texttt{002}              & 78.7              & 70.7              & 48.6 \\
            symmetric-\texttt{006}              & 77.5              & 63.9              & 38.2 \\
            symmetric-\texttt{014}              & 71.9              & 59.0              & 47.8 \\
            symmetric-\texttt{017}              & \textbf{88.0}     & 72.3              & 50.6 \\
            symmetric-\texttt{022}              & 85.9              & 73.5              & 59.4 \\
            symmetric-\texttt{052}              & 85.9              & 71.5              & 55.0 \\
            symmetric-\texttt{053}              & 84.3              & \textbf{76.7}     & 55.4 \\
            symmetric-\texttt{293}              & 85.5              & 74.3              & \textbf{59.8} \\
            symmetric-\texttt{323}              & 87.6              & 76.3              & 55.4 \\
            \hline
            10-scan PCA                         & 86.0              & 65.9              & 42.2 \\
            10-scan KDE                         & \textbf{94.0}     & \textbf{85.9}     & \textbf{71.5} \\
            \hline

            BU3D-FE  \cite{gerig2018morphable}  & 90.4              & 82.7              &  68.7 \\ 
            BFM '17 \cite{gerig2018morphable}   & \textbf{98.8}     & \textbf{98.0}     & \textbf{90.0} \\ 
            \hline
        \end{tabular}
        \caption{Face recognition results for images from the Multi-PIE database \cite{gross_multi-pie_2010}.  Each column represents the accuracy for a set of probe images with a common yaw angle given in the first row.  The second row gives the common ending of the IDs in the Multi-PIE dataset of the probe images with a given yaw angle.  The gallery is constructed from the images of all 249 identities with a yaw angle of $0^{\circ}$ (dataset IDs ending in \texttt{051\_16}).  Chance rate is $0.4$.  The 3DMMs in the second box (standard-full to symmetric-XYZ) were produced using the mean of the 2019 Basel Face Model \cite{gerig2018morphable}, while the 3DMMs in the third box (symmetric-\texttt{001} to symmetric-\texttt{323}) were produced using the 3D scans provided with the 2009 Basel Face Model \cite{paysan20093d}.  BFM '17 refers to the 2017 Basel Face Model \cite{gerig2018morphable}.}
        \label{tab:fr_experiment}
    \end{table}
    
    In our second experiment, we use the inverse rendering used above to perform face recognition, as outlined in \cite{schonborn2017markov, gerig2018morphable, blanz_face_2003}.  By reconstructing the shape and albedo latents from a gallery of reference images $\{f_1, \ldots, f_n\}$ (with one image per identity), we can obtain latents $(c_{s, i}, c_{a, i})$ for each reference image $f_i$.  Faces in a novel image $f_0$ are then identified by reconstructing shape and albedo latents $(c_{s, 0}, c_{a, 0})$ from said image and determining the reference image with the maximum cosine angle in the joint shape-albedo latent space, as in \cite{blanz_face_2003}.  We conduct face recognition on images from the CMU Multi-PIE database \cite{gross_multi-pie_2010}.  The results are presented in Table~\ref{tab:fr_experiment}.
    
    Table~\ref{tab:fr_experiment} illustrates that the 3DMMs with albedo kernels that combine RGB-space and physical-space distance information perform face recognition significantly more accurately on all image types than do 3DMMs with albedo kernels that only make use of one type of distance metric.  Furthermore, we may observe that the performance of the symmetric model is better on all image types than that of the BU3D-FE model \cite{gerig2018morphable}, a 3DMM built from 100 3D scans.  Table~\ref{tab:fr_experiment} also illustrates that 3DMMs defined using the mean of the 2019 Basel Face Model have better performance than those defined using individual face scans.  This is particularly true on images with a yaw angle over $15^{\circ}$, since as the yaw angle increases, the prior (in this case the 3DMM) plays a larger role in generating the reconstruction.

    The previously presented face recognition results relied on the mean of the 2019 Basel Face Model \cite{gerig2018morphable}.  The performances of the 3DMMs built using individual face scans (symmetric-\texttt{001} to symmetric-\texttt{323}) are also listed in Table~\ref{tab:fr_experiment}.  The performances of these 3DMMs are clearly significantly lower than that of the symmetric-full 3DMM.  However, by combining the information present in the 10 scans through our KDE approach, we can produce a new model that achieves performance comparable to that of the symmetric-full 3DMM.  To perform face recognition with this non-parametric model, we perform inference for each mixture component separately on both the probe image and each gallery image.  We then compute the cosine-angle in latent space between the probe reconstruction and all gallery reconstructions for each mixture component, and classify the probe image based on which 3DMM and gallery image yields the smallest cosine-angle.
    
    The performance of this mixture model on our face recognition task is listed in Table~\ref{tab:fr_experiment} as ``10-scan KDE''.  As Table~\ref{tab:fr_experiment} shows, this approach offers face recognition performance comparable to that achieved by the symmetric-full 3DMM, and outperforms the BU3D-FE model on all yaw levels, despite using only 10 scans.  To provide a more direct comparison between our novel KDE approach and PCA-based 3DMMs, we also produced a 3DMM by performing PCA with the 10 scans.  The face recognition performance of this 3DMM is listed in Table~\ref{tab:fr_experiment} as ``10-scan PCA''.  Table~\ref{tab:fr_experiment} demonstrates that this PCA-based 3DMM has far poorer face recognition performance than our KDE-based model.  In fact, the performance of the 10-scan PCA-based 3DMM is comparable to that of the 3DMMs produced from a single individual face scan (symmetric-\texttt{001} to symmetric-\texttt{323}).
    
\subsection{Learned Models}\label{sec:learnedmodels}

    \begin{figure}[H]
        \centering
        \begin{tabular}{|c|c|}
            \hline
            smooth albedo & smooth shape\\
            \hline
            \includegraphics[height=2cm]{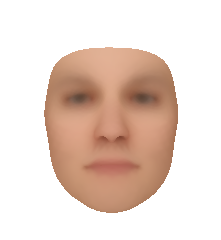}  \includegraphics[height=2cm]{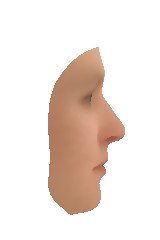} & \includegraphics[height=2cm]{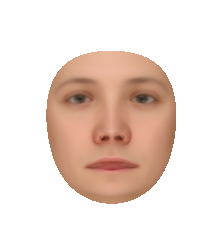}  \includegraphics[height=2cm]{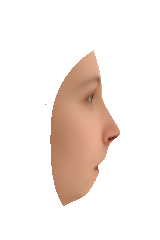}\\
            \hline
        \end{tabular}
        \caption{Simplified 3D face templates used to initialize the learning process used to construct the smooth-albedo-1, smooth-albedo-5, smooth-shape-1, and smooth-shape-5 models shown in Figures~\ref{fig:learned_samples} and ~\ref{fig:learned_pcs}.  These 3D templates were constructed by applying blur transformations to the mean of the 2019 Basel Face Model \cite{gerig2018morphable}.}
        \label{fig:simplified_scans}
    \end{figure}

    \begin{figure}[H]
        \centering
        \begin{tabular}{|l|c|c|c|}
            \hline
            standard-1 & \includegraphics[height=1.5cm]{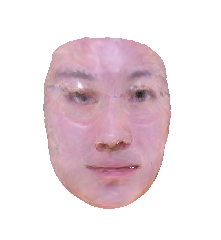} \includegraphics[height=1.5cm]{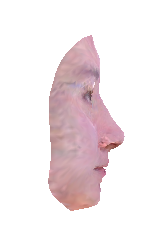} & \includegraphics[height=1.5cm]{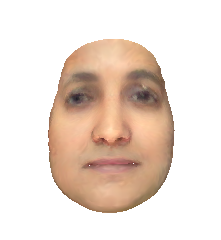}  \includegraphics[height=1.5cm]{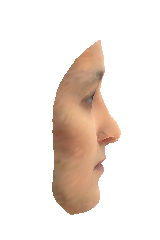} & \includegraphics[height=1.5cm]{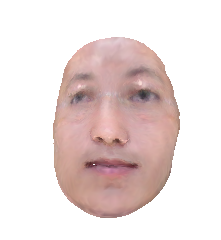}  \includegraphics[height=1.5cm]{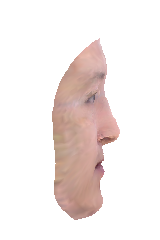} \\
            \hline
            standard-5 & \includegraphics[height=1.5cm]{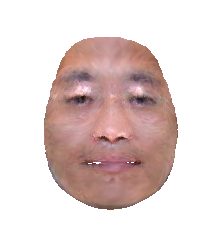} \includegraphics[height=1.5cm]{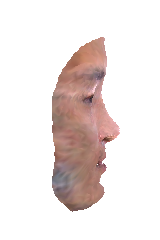} & \includegraphics[height=1.5cm]{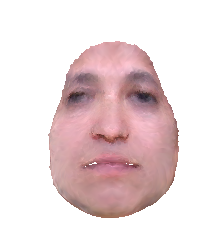}  \includegraphics[height=1.5cm]{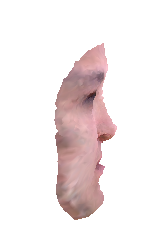} & \includegraphics[height=1.5cm]{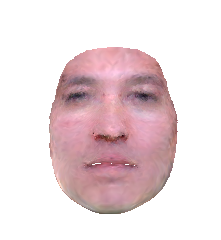}  \includegraphics[height=1.5cm]{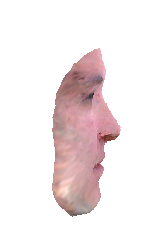} \\
            \hline
            symmetric-1 & \includegraphics[height=1.5cm]{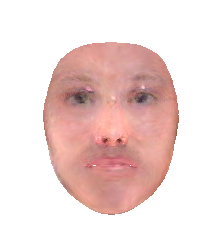} \includegraphics[height=1.5cm]{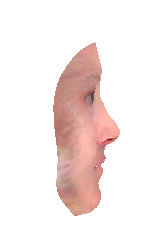} & \includegraphics[height=1.5cm]{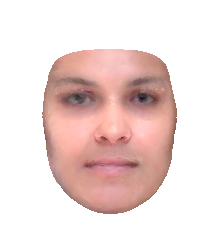} \includegraphics[height=1.5cm]{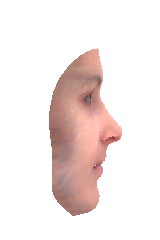} & \includegraphics[height=1.5cm]{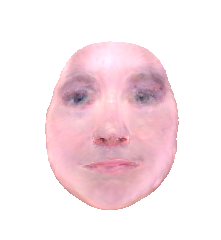} \includegraphics[height=1.5cm]{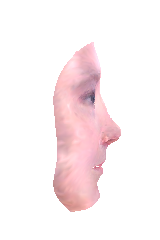} \\
            \hline
            symmetric-5 & \includegraphics[height=1.5cm]{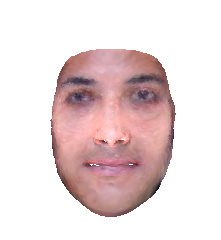} \includegraphics[height=1.5cm]{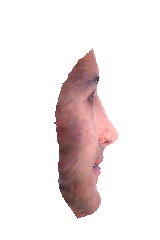} & \includegraphics[height=1.5cm]{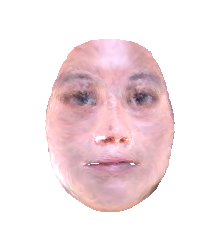} \includegraphics[height=1.5cm]{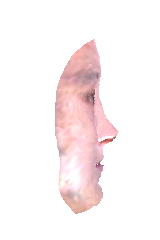} & \includegraphics[height=1.5cm]{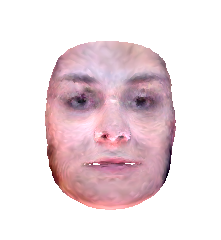} \includegraphics[height=1.5cm]{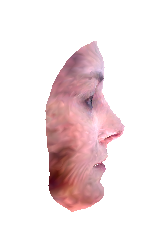} \\
            \hline
            smooth-albedo-1 & \includegraphics[height=1.5cm]{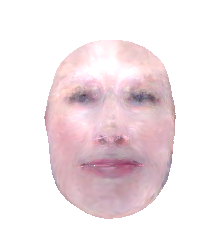} \includegraphics[height=1.5cm]{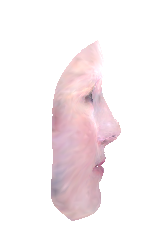} &  \includegraphics[height=1.5cm]{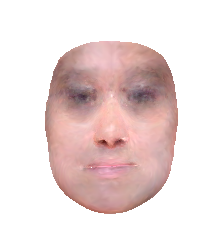} \includegraphics[height=1.5cm]{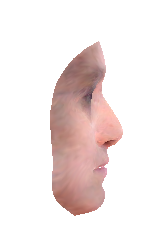} & \includegraphics[height=1.5cm]{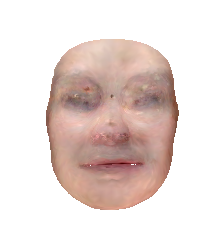} \includegraphics[height=1.5cm]{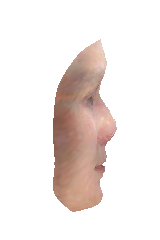} \\
            \hline
            smooth-albedo-5 &  \includegraphics[height=1.5cm]{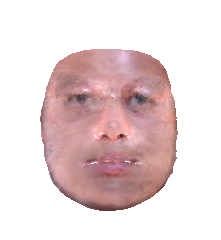} \includegraphics[height=1.5cm]{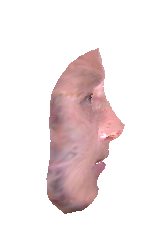} & \includegraphics[height=1.5cm]{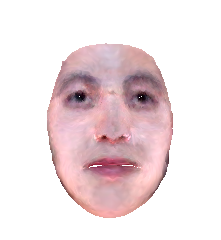} \includegraphics[height=1.5cm]{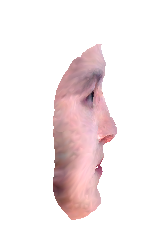} & \includegraphics[height=1.5cm]{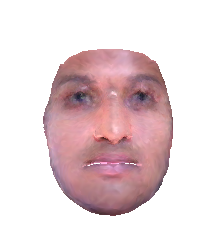} \includegraphics[height=1.5cm]{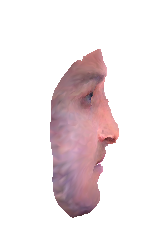} \\
            \hline
            smooth-shape-1 &  \includegraphics[height=1.5cm]{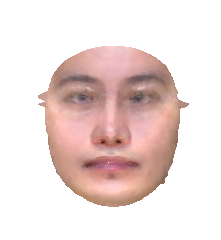} \includegraphics[height=1.5cm]{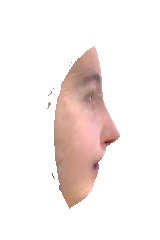} & \includegraphics[height=1.5cm]{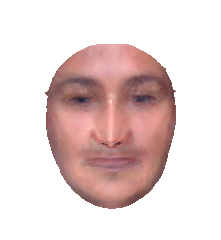} \includegraphics[height=1.5cm]{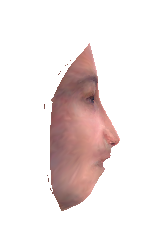} & \includegraphics[height=1.5cm]{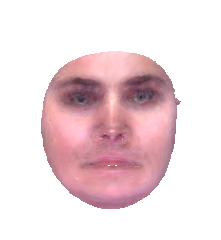} \includegraphics[height=1.5cm]{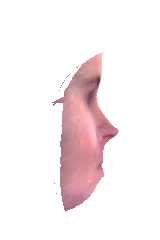} \\
            \hline
            smooth-shape-5 &  \includegraphics[height=1.5cm]{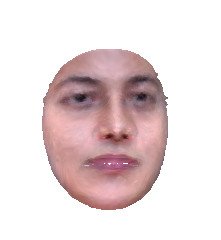} \includegraphics[height=1.5cm]{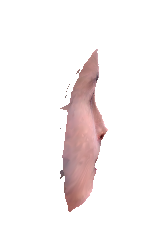} & \includegraphics[height=1.5cm]{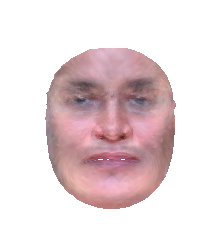} \includegraphics[height=1.5cm]{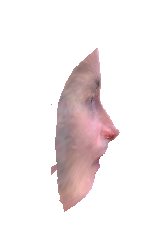} & \includegraphics[height=1.5cm]{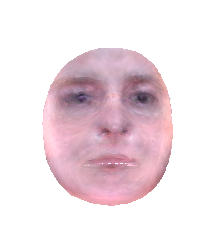} \includegraphics[height=1.5cm]{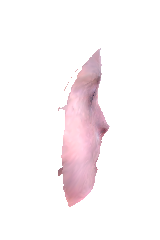} \\
            \hline
            LeMoMo & \includegraphics[height=1.5cm]{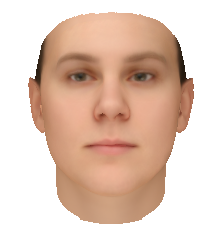} \includegraphics[height=1.5cm]{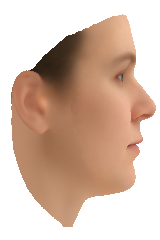} & \includegraphics[height=1.5cm]{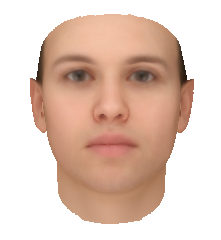} \includegraphics[height=1.5cm]{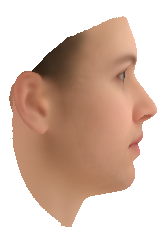} & \includegraphics[height=1.5cm]{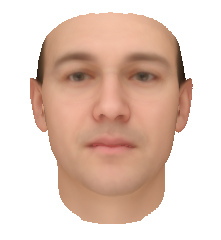} \includegraphics[height=1.5cm]{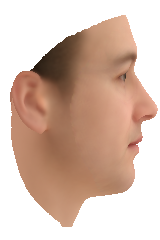} \\
            \hline
        \end{tabular}
        \caption{Samples from the models learned through the learning process outlined in Section~\ref{sec:learning}, using as initial 3DMMs the standard-full (standard-1 and standard-5) and symmetric-full (symmetric-1 and symmetric-5) 3DMMs, as well as analogues of the symmetric-full 3DMMs built using initial scans with simplified albedo (smooth-albedo-1 and smooth-albedo-5) and simplified shape (smooth-shape-1 and smooth-shape-5), and the LeMoMo model developed by Tewari et al. \cite{tewari2020learning}.  Learning was performed for one iteration (standard-1, symmetric-1, smooth-albedo-1, and smooth-shape-1) or for five iterations (standard-5, symmetric-5, smooth-albedo-5, smooth-shape-5).}
        \label{fig:learned_samples}
    \end{figure}
    
    \begin{figure}[H]
        \centering
        \begin{tabular}{|l|c|c|c|c|c|}
            \hline
            & mean & $+$albedo & $-$albedo & $+$shape & $-$shape\\
            \hline
            standard-1 & \includegraphics[height=1.5cm]{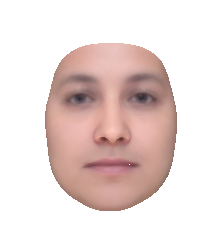} & \includegraphics[height=1.5cm]{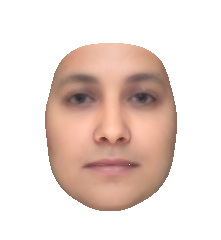} & \includegraphics[height=1.5cm]{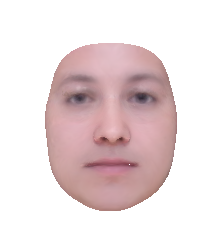} & \includegraphics[height=1.5cm]{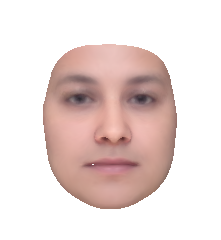} &
            \includegraphics[height=1.5cm]{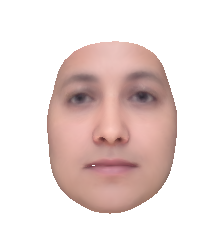} \\
            \hline
            standard-5 & \includegraphics[height=1.5cm]{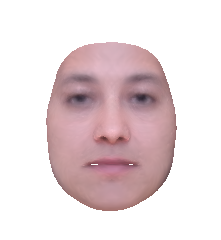} & \includegraphics[height=1.5cm]{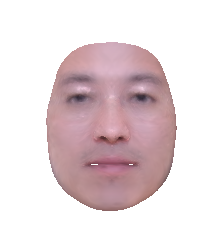} & \includegraphics[height=1.5cm]{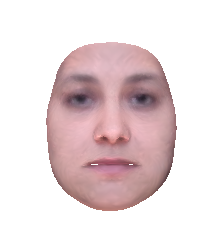} & \includegraphics[height=1.5cm]{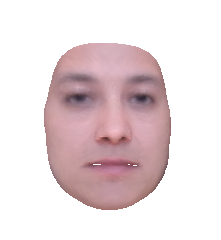} &
            \includegraphics[height=1.5cm]{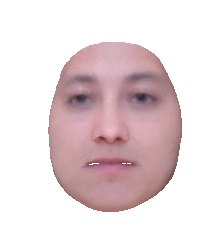} \\
            \hline
            symmetric-1 & \includegraphics[height=1.5cm]{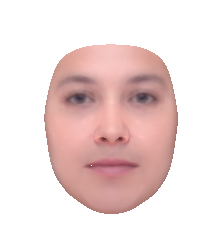} & \includegraphics[height=1.5cm]{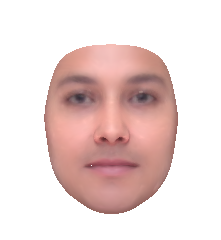} & \includegraphics[height=1.5cm]{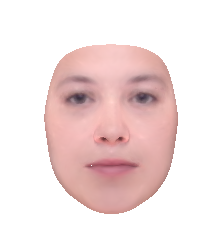} & \includegraphics[height=1.5cm]{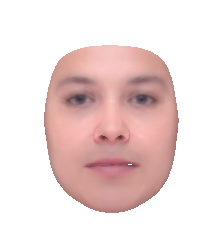} &
            \includegraphics[height=1.5cm]{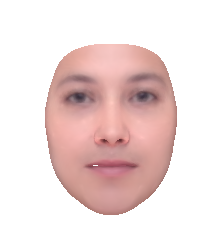} \\
            \hline
            symmetric-5 & \includegraphics[height=1.5cm]{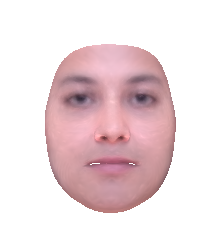} & \includegraphics[height=1.5cm]{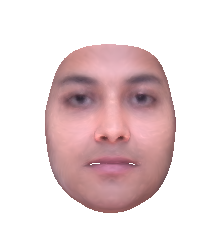} & \includegraphics[height=1.5cm]{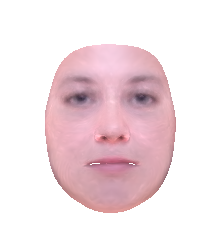} & \includegraphics[height=1.5cm]{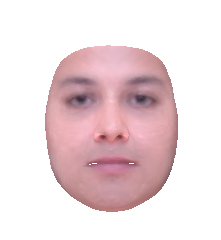} &
            \includegraphics[height=1.5cm]{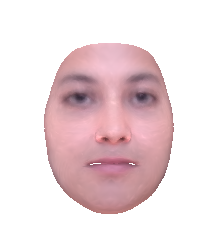} \\
            \hline
            smooth-albedo-1 & \includegraphics[height=1.5cm]{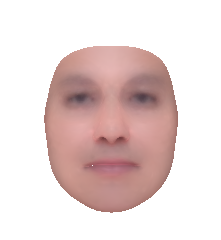} & \includegraphics[height=1.5cm]{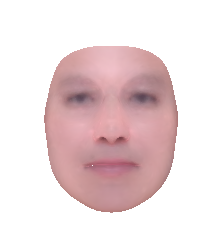} & \includegraphics[height=1.5cm]{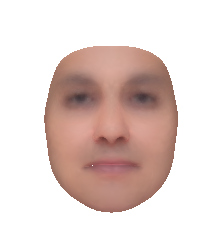} & \includegraphics[height=1.5cm]{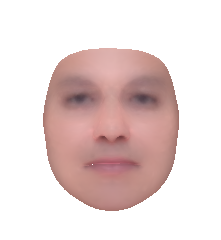} &
            \includegraphics[height=1.5cm]{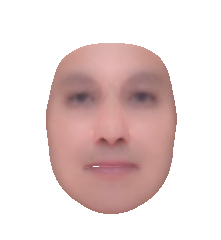} \\
            \hline
            smooth-albedo-5 & \includegraphics[height=1.5cm]{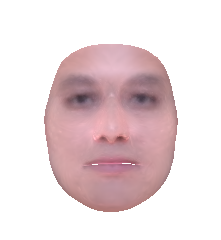} & \includegraphics[height=1.5cm]{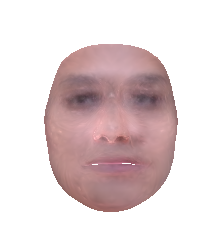} & \includegraphics[height=1.5cm]{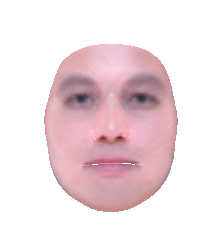} & \includegraphics[height=1.5cm]{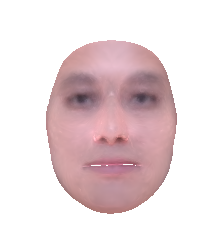} &
            \includegraphics[height=1.5cm]{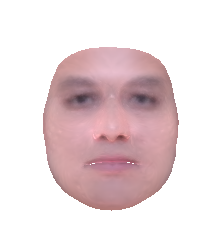} \\
            \hline
            smooth-shape-1 & \includegraphics[height=1.5cm]{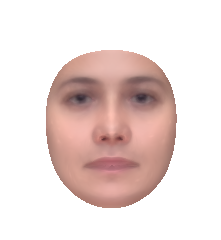} & \includegraphics[height=1.5cm]{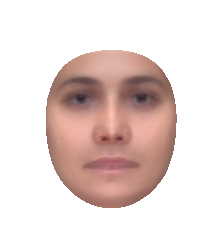} & \includegraphics[height=1.5cm]{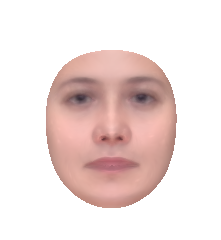} & \includegraphics[height=1.5cm]{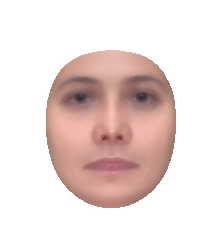} &
            \includegraphics[height=1.5cm]{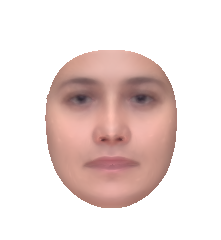} \\
            \hline
            smooth-shape-5 & \includegraphics[height=1.5cm]{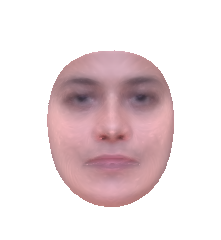} & \includegraphics[height=1.5cm]{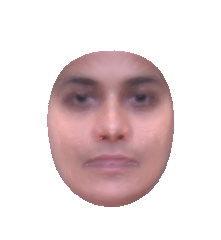} & \includegraphics[height=1.5cm]{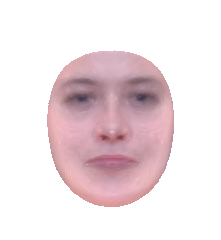} & \includegraphics[height=1.5cm]{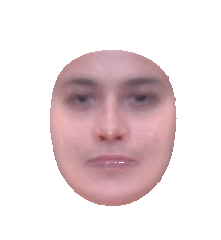} &
            \includegraphics[height=1.5cm]{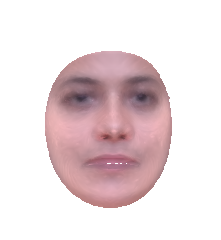} \\
            \hline
            LeMoMo & \includegraphics[height=1.5cm]{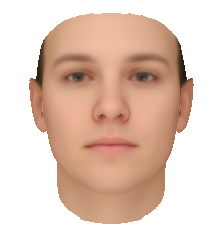} & \includegraphics[height=1.5cm]{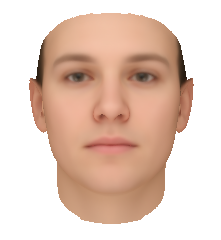} & \includegraphics[height=1.5cm]{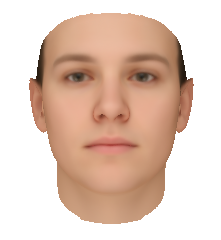} & \includegraphics[height=1.5cm]{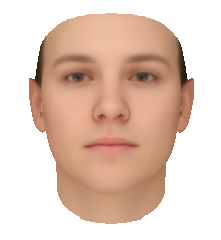} & \includegraphics[height=1.5cm]{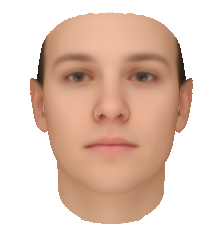} \\
            \hline
        \end{tabular}
        \caption{The mean, as well as the mean offset by $\pm 1$ times the first principal component of the albedo (``$+$albedo'', ``$-$albedo'') or shape (``$+$shape'', ``$-$shape'') models of the standard-1, standard-5, symmetric-1, symmetric-5, smooth-albedo-1, smooth-albedo-5, smooth-shape-1, and smooth-shape-5 models, and the LeMoMo model developed by Tewari et al. \cite{tewari2020learning}.}
        \label{fig:learned_pcs}
    \end{figure}
    
    Despite performing well on face recognition tasks, samples from our single-scan 3DMMs are nevertheless highly non-naturalistic, as shown in Figure~\ref{fig:kernel_illustrations}.  Using the learning approach outlined in Section~\ref{sec:learning}, we augmented our standard-full and symmetric-full 3DMMs using 200 images from the Multi-PIE dataset \cite{gross_multi-pie_2010}.  We used images of 50 distinct individuals shown in a frontal perspective, from a $15^{\circ}$ angle, from a $30^{\circ}$ angle, and from $45^{\circ}$ angle.  However, no identity or pose annotations were used; our learning algorithm treated each image as if it was an image of a novel individual in an unknown pose.  We augmented these 200 images by adding alternate versions of each image that were flipped left-to-right, so our learning algorithm used 400 images total.
    
    We ran five iterations of our wake-sleep-like procedure, creating ten new 3DMMs total; here we show only four of them, namely the 3DMMs produced after one iteration, and the 3DMMs produced after all five iterations.  We name these 3DMMs ``standard-1'', ``standard-5'', ``symmetric-1'', and ``symmetric-5''.  Random samples from these 3DMMs shown in frontal and side views are shown in Figure~\ref{fig:learned_samples}.  
    
    We additionally wanted to see if our learning procedure could reconstruct the mean face even when initialized with a simplified face scan.  To do so, we applied a blur filter to (separately) the shape and albedo of the mean of the 2019 Basel Face Model \cite{gerig2018morphable}, resulting in the simplified meshes shown in Figure~\ref{fig:simplified_scans}, and constructed analogues of the symmetric-full models using these simplified meshes, yielding two new 3DMMs.  We then reran our learning algorithm using these new 3DMMs as initializations, yielding 10 new learned 3DMMs; we again show only four of them, namely those produced after one iteration and after all five iterations.  We name the 3DMMs produced using a simplified albedo ``smooth-albedo-1'' and ``smooth-albedo-5'', and those produced using a simplified shape ``smooth-shape-1'' and ``smooth-shape-5''.  In Figure~\ref{fig:learned_samples}, we also show random samples (in frontal and side views) from the analogues of standard-1, standard-5, symmetric-1, and symmetric-5 3DMMs produced using the simplified initial meshes.  In addition to showing samples from these distributions, we can also examine their means (that is, the means of the learned distributions, not the initial scan used to build the models) and their first principal components.  In Figure~\ref{fig:learned_pcs}, we show the means of the standard-1, standard-5, symmetric-1, symmetric-5, smooth-albedo-1, smooth-albedo-5, smooth-shape-1, and smooth-shape-5 3DMMs, in front and side views.  Additionally, Figure~\ref{fig:learned_pcs} also shows these mean altered by adding their respective 3DMMs' first shape or albedo principal components.
    
\begin{figure}[H]
        \begin{center}
            \includegraphics[width=0.95\linewidth]{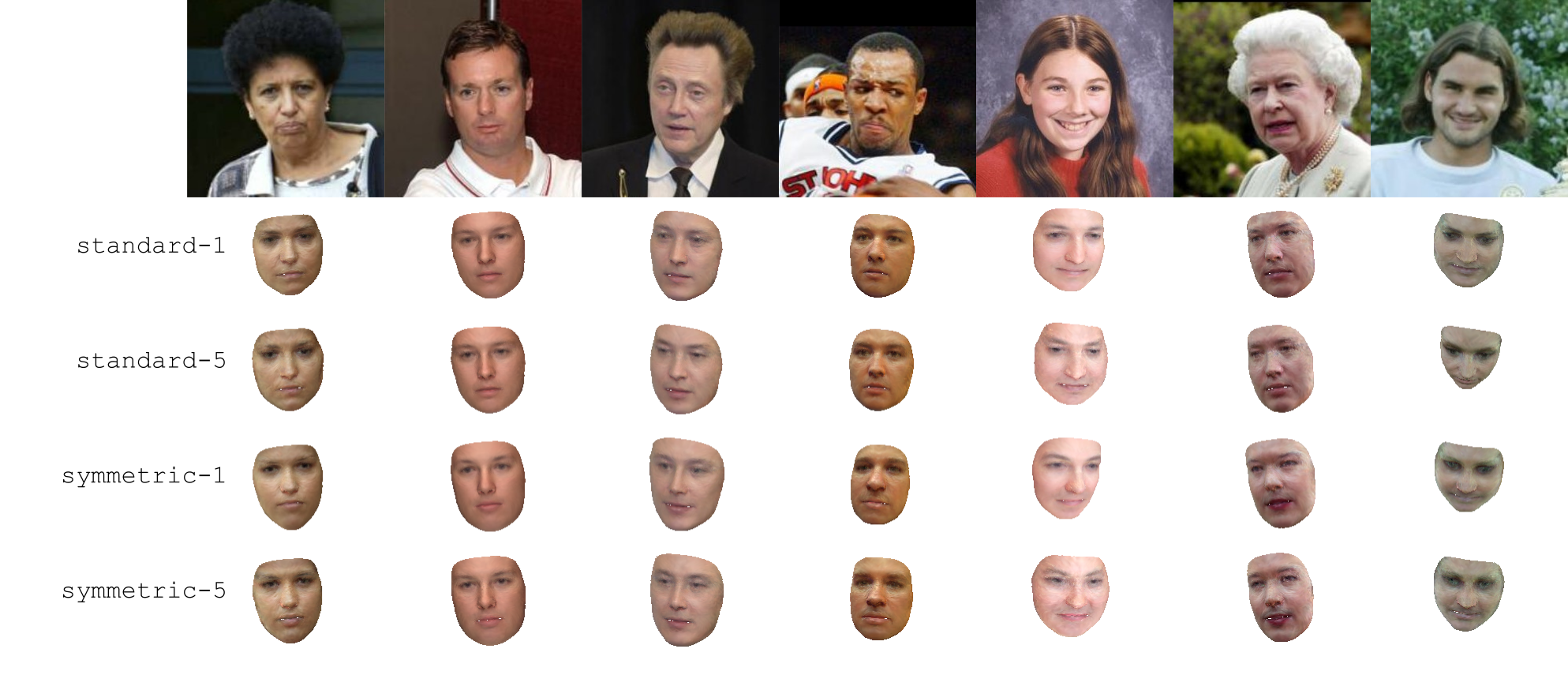}
        \end{center}
        \caption{Reconstructions produced from natural images using several of our learned 3DMMs.  As in Figure~\ref{fig:lfw_results}, the first row shows the images used as input, and the subsequent rows show 3D reconstructions produced using our 3DMMs.}
        \label{fig:learned_lfw_results}
    \end{figure}

    Finally, Figure~\ref{fig:learned_lfw_results} shows qualitative 3D face reconstructions produced from images from the Labeled Faces in the Wild dataset \cite{huang2008labeled} using the standard-1, standard-5, symmetric-1, and symmetric-5 models, and the inference method of \cite{schonborn2017markov}.  This directly mirrors Figure~\ref{fig:lfw_results}, with the only difference being that the reconstructions are produced using different 3DMMs.  While the reconstructions in Figure~\ref{fig:learned_lfw_results} are not visually much better than those in Figure~\ref{fig:lfw_results}---likely because the increased realism of the learned models comes with a reduction in flexibility---they nevertheless demonstrate that our learned models can likewise be used in basic inverse graphics settings.
    
    While the resulting face 3DMMs are still non-naturalistic in some ways they seem a clear improvement over our initial 3DMMs, as can be seen by comparing Figure~\ref{fig:learned_samples} with Figure~\ref{fig:kernel_illustrations}.  For instance, the standard-1 and standard-5 models have clearly learned naturalistic face tones as well as approximate facial symmetry, which were lacking from the initial standard-full 3DMM.  Figure~\ref{fig:learned_pcs} demonstrates that although the smooth-shape-1 and smooth-shape-5 models do not appear to be able to learn a realistic face shape given a highly unrealistic template mesh, the smooth-albedo-1 and smooth-albedo-5 do appear to be able to learn a realistic mean face albedo even when the initial template mesh possesses a non-naturalistic albedo.  While our learned 3DMMs show a significant visual improvement over the standard-full and symmetric-full 3DMMs, quantitatively demonstrating an improvement has proven difficult.  Face recognition performance as measured in Table~\ref{tab:fr_experiment} is significantly lower with the learned models than with our initial standard-full and symmetric-full 3DMMs, and shape specificity and generalization (as shown in Figure~\ref{fig:spec_gen}) are also far lower, while albedo specificity and generalization are comparable or somewhat poorer.  This may be at least partially due to artifacts of the learning process; in particular, for the face recognition results, the number of principal components is much lower in the learned models because it is limited by the number of reconstructions which pass the quality control process, while the vastly lower shape specificity might be a byproduct of the alignment process applied during learning.

\subsection{Specificity, Generalization, and Compactness}\label{sec:specgen}

\begin{figure}[H]
        \begin{center}
            \includegraphics[width=0.95\linewidth]{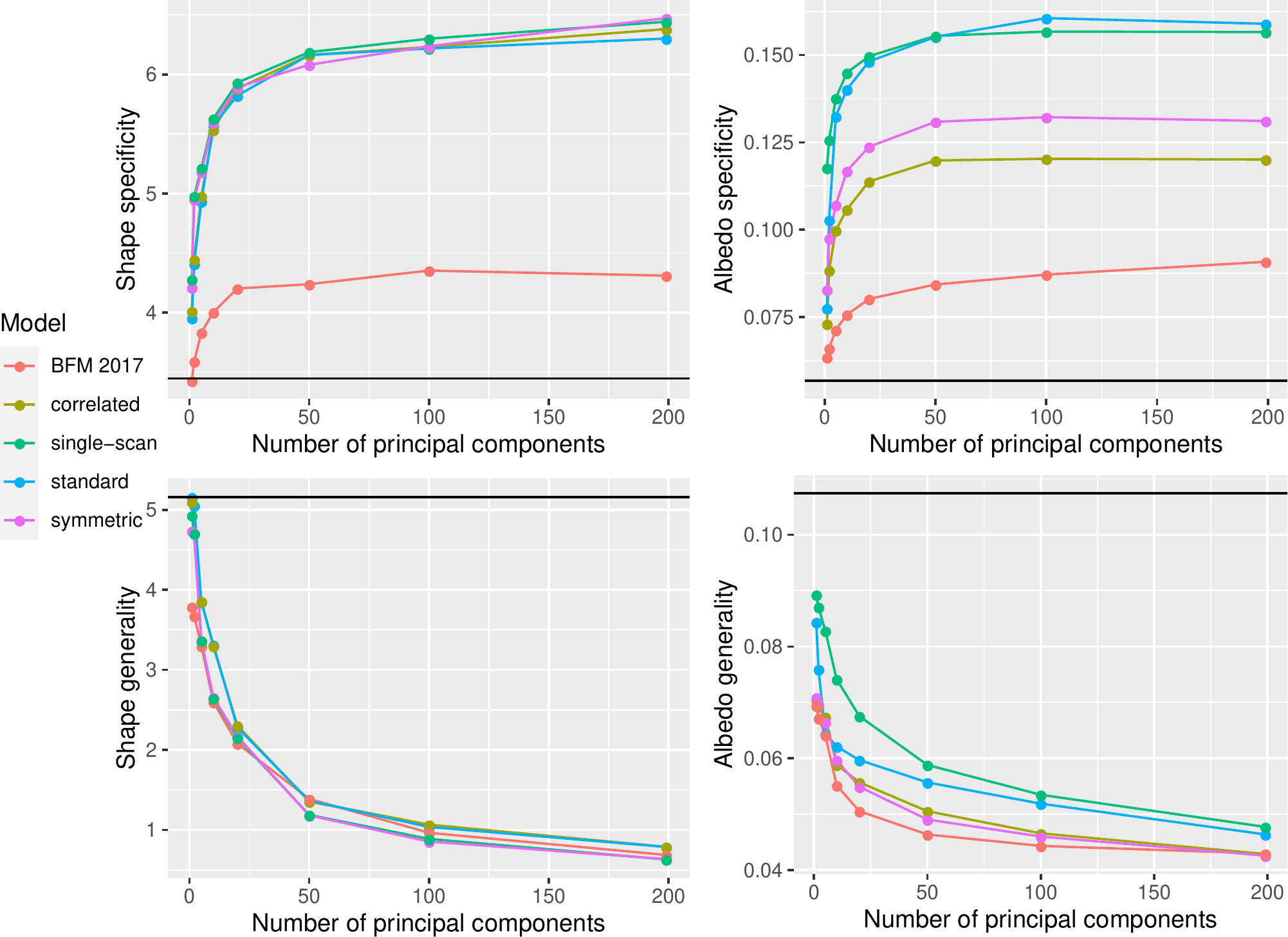}
        \end{center}
        \caption{A plot of the specificity, generalization, and compactness \cite{styner2003evaluation} of our 3DMMs' relative to the 2017 Basel Face Model \cite{gerig2018morphable}.  ``standard'', ``correlated'', and ``symmetric'' refer to versions of the standard-full, correlated-full, and symmetric-full models built using the mean of the 2017 Basel Face Model, while the ``single-scan'' results are an average of the performance of the various symmetric-$x$ models.  The scans included with the 2009 Basel Face Model \cite{paysan20093d} were used as a dataset; for the symmetric-$x$ 3DMMs, the scan used to construct the 3DMM was excluded.  See Section~\ref{sec:specgen} for more details.}
        \label{fig:spec_gen}
    \end{figure}

    Figure~\ref{fig:spec_gen} shows plots of the specificity, generalization, and compactness \cite{styner2003evaluation} of our 3DMMs and the 2017 Basel Face Model \cite{gerig2018morphable}; specifically, it shows the specificity and generalization of the shape and albedo models of each 3DMM as a function of the number of principal components included.  We compare the 2017 Basel Face Model (``BFM 2017'') and versions of the standard-full (``standard''), symmetric-full (``symmetric''), and correlated-full (``correlated'') models built using the mean of the 2017 Basel Face Model as template.  The correlated-full model, presented in the supplementary material, is analogous to the symmetric-full model but lacks bilateral symmetry.  We use as our dataset the ten scans provided with the 2009 Basel Face Model \cite{paysan20093d}.  We also include the symmetric-$x$ models, where $x$ is a scan ID number (\texttt{001}, \texttt{002}, \texttt{006}, \texttt{014}, \texttt{017}, \texttt{022}, \texttt{052}, \texttt{053}, \texttt{293}, or \texttt{323}); for these models we exclude the scan used to build the model.  We report results averaged across the symmetric-$x$ models as ``single-scan''.  We measure specificity and generalization using 1, 2, 5, 10, 20, 50, 100, and all (199) principal components.  We indicate the specificity and generalization of the mean of the 2017 Basel Face Model, considered as a 3DMM with zero principal components, with a black line.
    
    We may observe that, for all numbers of principal components, the generalization of our 3DMMs' shape models is comparable to that of the 2017 Basel Face Model, while the generalization of our 3DMMs' albedo models is in fact superior to that of the 2017 Basel Face Model.  The specificity of our 3DMMs' shape and albedo models, is, of course, inferior to that of the 2017 Basel Face Model.  This is unavoidable as our models' were constructed using far less data than the 2017 Basel Face Model.  We may additionally observe that our single-scan models perform comparably to the standard-full model across all conditions.

\subsection{Registration Tasks}\label{subsec:registration}

    Registration is another task for which 3DMMs may be used.  In this task we wish to transform an arbitrary face mesh into a mesh with a given topology while preserving the face as closely as possible.  Prior work has nearly exclusively relied on shape information to compute such a transformation \cite{egger20203d}.  However, albedo information also provides important constraints on face registration.  For instance, the eyebrows and the pupils of the eyes are almost entirely defined by albedo.
    
    \begin{figure}[H]
        \begin{center}
            \begin{tabular}{|c|c|c|}
                \hline
                shape only & shape and albedo & BFM'09\\
                \hline
                \includegraphics[width=0.25\linewidth]{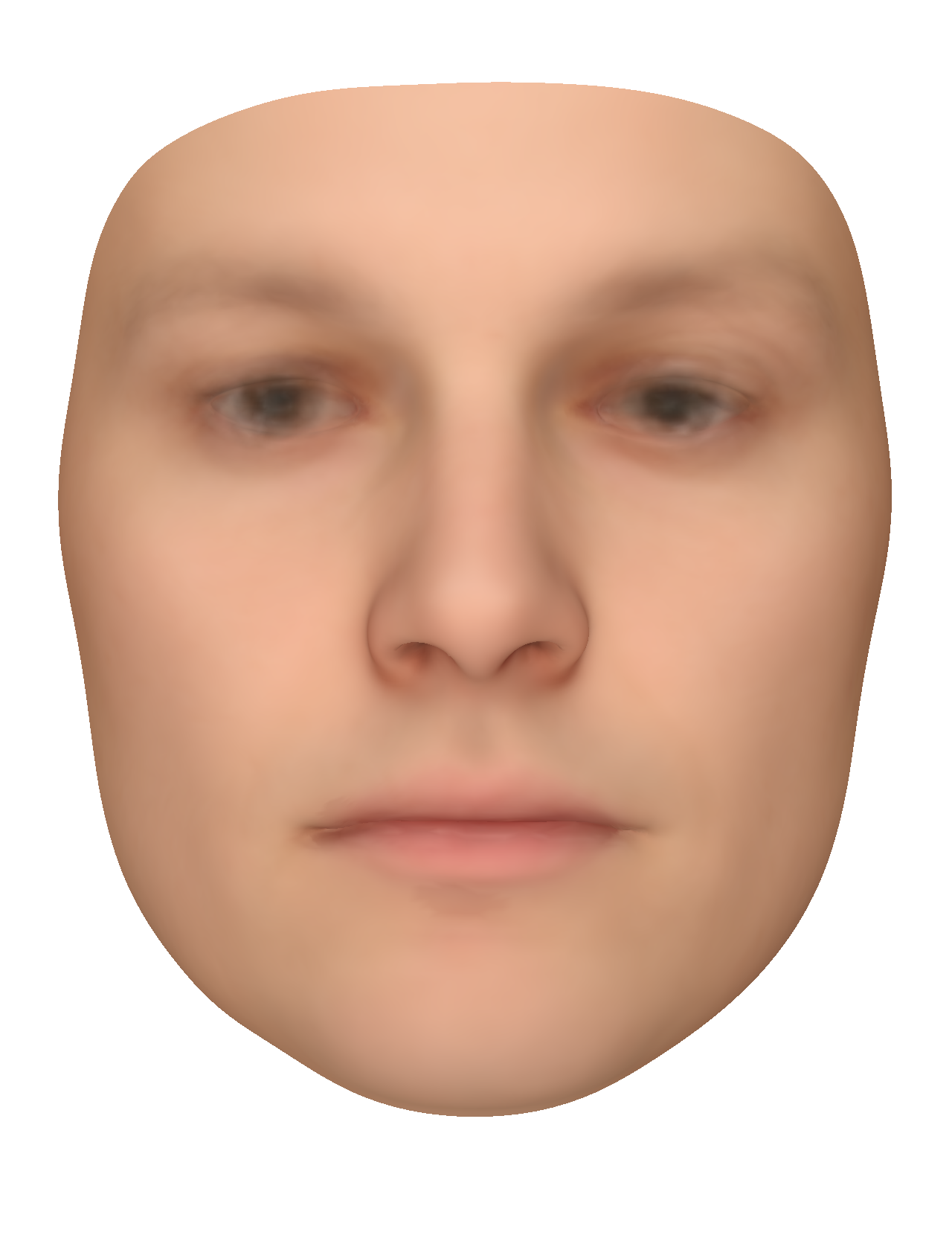} &
                \includegraphics[width=0.25\linewidth]{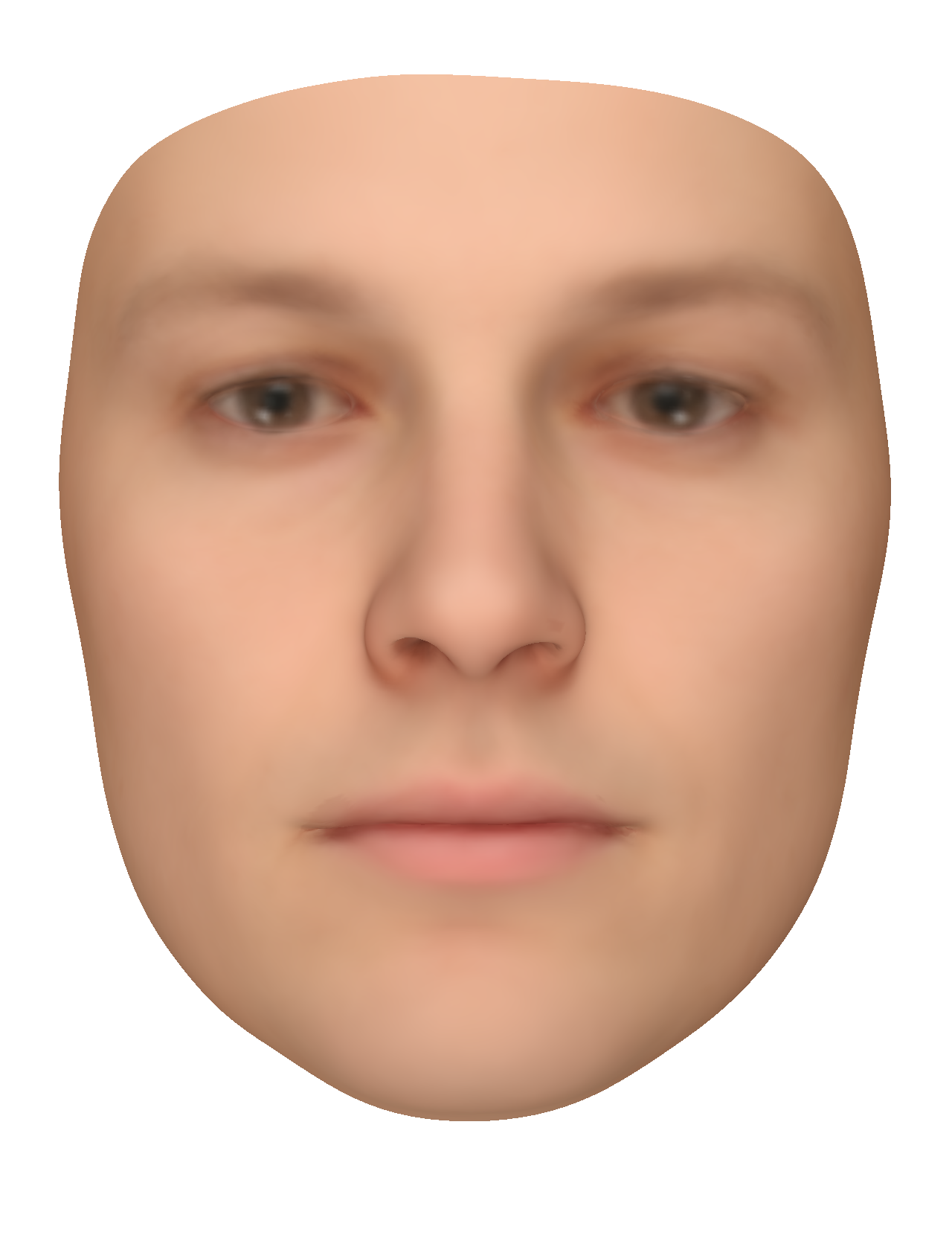} &
                \includegraphics[width=0.25\linewidth]{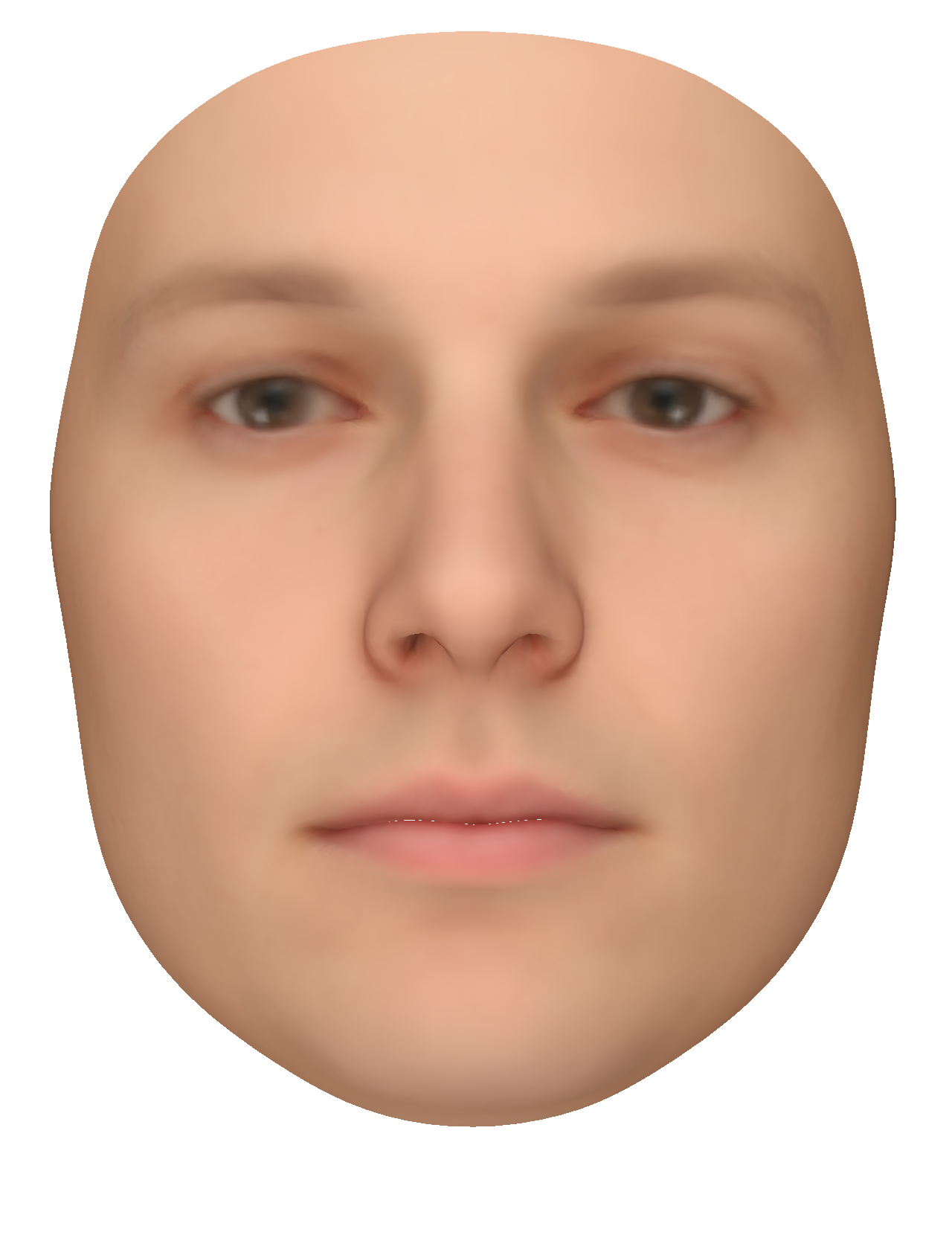}\\
                \includegraphics[width=0.25\linewidth]{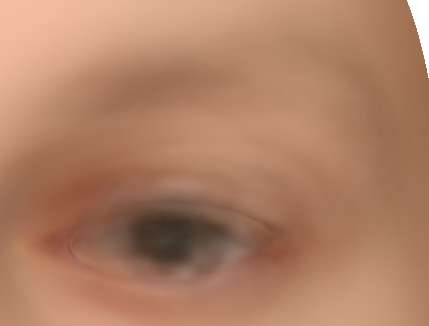} &
                \includegraphics[width=0.25\linewidth]{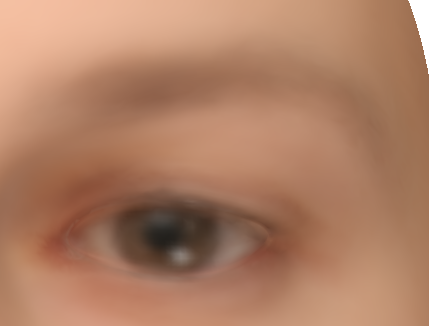} &
                \includegraphics[width=0.25\linewidth]{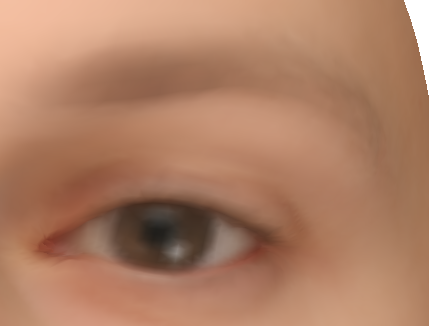}\\
                \hline
            \end{tabular}
        \end{center}
        \caption{The average of the registration results produced by the MCMC methods using both shape and albedo information (``shape and albedo'') or shape information only (``shape only'') on all ten scans, along with the average of the corresponding registered meshes produced in the construction (strongly reliant on manual landmark annotations) of the 2009 Basel Face Model \cite{paysan20093d} (``BFM'09'').  Close-ups of the left eye and eyebrow are provided, illustrating that the eyebrows and the pupils of the eyes are far less clearly defined in the shape-only condition.}
        \label{fig:registration_albedo}
    \end{figure}
    
    To perform registration tasks with our 3DMMs, we adapted the inverse rendering approach of \cite{schonborn2017markov} to minimize the chamfer distance between the model mesh and the target mesh while simultaneously minimizing the pixel error between the rendered model instance and the rendered target mesh.  We achieve this by combining an image-based reconstruction likelihood, which constrains 2D appearance, with a shape-based likelihood, which enforces 3D shape consistency as measured by chamfer distance.  This minimizes shape distance and induces albedo consistency while establishing correspondence with the topology of our 3DMM's template.  While both those ideas are often applied in isolation, they are rarely combined in registration tasks (or only combined as post-processing).  We roughly align the meshes to initialize the pose, but, unlike typical approaches, do not use landmarks during registration.  Instead, the location of facial features is constrained by the albedo component of the evaluation.  As post-processing we eliminate any net translation using the method of \cite{umeyama_least-squares_1991} and set each vertex's albedo by projecting vertex normals onto the scan as in \cite{gerig2018morphable}.
    
    This process enables us to make use of both shape and albedo information in registration.  We compare the result of doing so with the analogous registration result produced using only shape information in our MCMC process.  We apply both registration methods to the unprocessed meshes for face scans \texttt{001}, \texttt{002}, \texttt{006}, \texttt{014}, \texttt{017}, \texttt{022}, \texttt{052}, \texttt{053}, \texttt{293}, and \texttt{323}.  To do so we use the standard-full 3DMM with the mean of the highest point-count version of the 2019 Basel Face Model \cite{gerig2018morphable} as reference.  To evaluate our registration we build a 3DMM from the registration results using PCA.
    
    We compare these results with the registration used by the 2009 Basel Face Model \cite{paysan20093d}, which used shape information along with manual landmark annotations.  Figure~\ref{fig:registration_albedo} demonstrates that by using shape and albedo information our registration process produces a sharp and stable albedo reconstruction whose quality is comparable to that of the 2009 Basel Face Model's registration, and far superior to that produced using shape information alone.  This performance is impressive, since the 2009 Basel Face Model heavily relied on human-provided landmark annotations in its registration pipeline, whereas our approach requires no annotations.
    
    The supplementary material contains a quantitative assessment of our shape registration performance, and shows that including albedo information in registration slightly increases the shape error.  This is unsurprising, as the shape-only reconstruction is optimized to produce the lowest shape error possible, and the shape and albedo reconstruction by definition cannot have less than the minimum shape error.  However, as Figure~\ref{fig:registration_albedo} demonstrates, the shape and albedo reconstruction has far higher quality overall.

\subsection{Constructing 3DMMs for Other Objects}\label{subsec:other_objects}

    \begin{figure}[H]
        \begin{center}
            \includegraphics[width=0.95\linewidth]{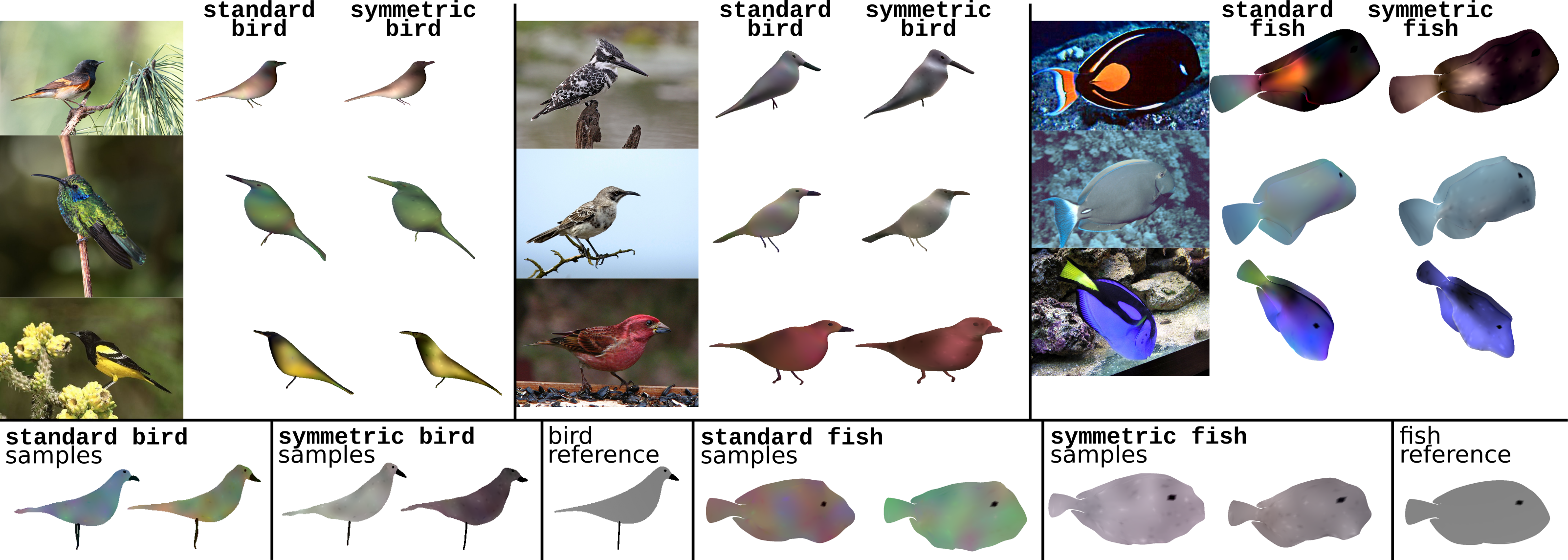}
        \end{center}
        \caption{On the upper left and middle: the reconstructions produced by the standard and symmetric bird models on six images taken from the Caltech-UCSD Birds 200 dataset \cite{wah_caltech-ucsd_2011}.  On the upper right: the reconstructions produced by the standard and symmetric fish models on three public-domain images taken from Wikipedia.  On the bottom: samples from the standard and symmetric bird and fish models, shown in side views, as well as the reference meshes used to build these 3DMMs.}
        \label{fig:bird_fish_lfw}
    \end{figure}

    We have thus far focused on 3DMM for faces; we now demonstrate that analogous methods can be used to build 3DMMs for other object categories.  Specifically, we construct single-scan 3DMMs for fish and birds using as references synthetic meshes with simple manual coloring.\footnote{Our bird and fish reference meshes are obtained from, respectively, \href{https://www.blendswap.com/blend/11752}{https://www.blendswap.com/blend/11752} and \href{https://www.turbosquid.com/3d-models/free-tail-animation-3d-model/368484}{https://www.turbosquid.com/3d-models/free-tail-animation-3d-model/368484}}  These meshes are simple artistic models and were constructed without 3D scanning.  We can build 3DMMs from each of these references using the same kernels as used in standard-full and symmetric-full, i.e. $K_s$ and $K_a$ in the first case, and $K_s^{\texttt{sym}}$ and $K_a^{\texttt{sym}}$ in the second.  This produces two new 3DMMs for each mesh, which we term the standard and symmetric models for each object category.  As our reference meshes lack many details that 3D scans possess, the performance of these 3DMMs is likely much lower than that of single-scan 3DMMs built from 3D scans.  Results with additional fish and bird 3DMMs produced with the $K_{a, \texttt{xyz}}$ and $K_{a, \texttt{xyz}}^{\texttt{sym}}$ albedo kernels are presented in the supplementary material.
    
    We seek to model a wide range of birds, but restrict ourselves to simple standing poses.  We restrict ourselves to the \textit{Acanthurus} genus of fish, which possesses a wide range of color variability but lack the fine details (such as scales) that many other fish possess.  In Figure~\ref{fig:bird_fish_lfw}, we show qualitative reconstruction results along with samples from our bird and fish models and the reference meshes used to construct them.  While these reconstructions are not as accurate as those in Figure~\ref{fig:lfw_results}, they do capture some rough features.  We suggest that three main factors make birds a more difficult object category than faces: birds have a much more complex albedo, including high-frequency components that our models capture poorly; birds have a well-defined silhouette, whereas faces have somewhat arbitrary boundaries; and color-correlation, while beneficial in modeling faces, impedes the symmetric bird model's ability to model birds.  Our standard model's performance on fish seems somewhat better, likely due to the lack of high-frequency components.  The symmetric model does much more poorly on fish, likely because the correlation of its color channels impedes its ability to model the regional color variation of fish.  It is important to keep in mind, however, that our method's performance is not directly comparable to that of other, less data-efficient approaches.%
    
\section{Conclusion}\label{sec:discussion}

    Our research demonstrates that we can build a simple 3DMM from a single template through the application of Gaussian process-based deformations. Although the result is of lower-quality than 3DMMs produced from high quality 3D scans, our simple models can still be used in many contexts where hand-produced 3DMMs have previously been required, and can be constructed using far less data and far simpler pipelines.  We demonstrate a preliminary unsupervised learning method for a 3DMM of faces from 2D images based solely on a single template without any supervision. For object categories where the number of available scans is extremely limited or where dense correspondence between scans cannot be easily obtained, this procedure thus offers a promising method for building 3DMMs.  Additionally, our results demonstrate the high value of fully integrating albedo into the 3DMM pipeline, and show that this can be done by combining covariance kernels which produce spatially continuous deformations with kernels that produce color-space-continuous deformations.  In addition to the results demonstrated in this paper, we believe our method can be highly beneficial in addressing dataset bias, a limitation of all currently available 3DMMs.
    
    In addition to its relevance in a computer vision context, our paper further demonstrates that a statistical model of faces can be learned from a initial simple template and limited unsupervised 2D data similar to what a human infant has access to.  This was motivated by interest in computationally modeling the cognitive development of human face perception in infants, and we hope that in the future our approach may inspire novel computational models of the development of human face perception.

\section*{Funding} 

    This work was funded by the DARPA Learning with Less Labels (LwLL) program (Contract No: FA8750-19-C-1001), the DARPA Machine Common Sense (MCS) program (Award ID: 030523-00001) and by the Center for Brains, Minds and Machines (CBMM) (NSF STC award CCF-1231216). B. Egger was supported by a PostDoc Mobility Grant, Swiss National Science Foundation P400P2\_191110.
    
\section*{Conflicting Interests} 
    The authors have no competing interests to declare that are relevant to the content of this article.

\clearpage
\bibliography{egbib}

\clearpage

\begin{appendices}
\section{Color-Correlated Asymmetric 3DMMs}\label{sec:correlated}

    In our main paper we build 3DMMs using Gaussian processes that include symmetry and color-channel correlation heuristics.  To assess the effects of these heuristics individually, we can also build 3DMMs that include only one of these heuristics.  Specifically, we experimented with constructing 3DMMs whose albedo models have correlated color channels but which lack symmetry.  This enables us to compare the relative importance in an analysis-by-synthesis setting of the symmetry and color-correlation heuristics of our symmetric 3DMMs.
    
    Our main paper defined albedo kernels $K_{a, \texttt{xyz}}^{\texttt{cor}}$ and $K_{a, \texttt{rgb}}^{\texttt{sym}}$.  These kernels possess a color-channel correlation heuristic but lack an explicit symmetry heuristic ($K_{a, \texttt{rgb}}^{\texttt{sym}}$ is symmetric, but this is only because we assume that the reference face is symmetric).  We may average these kernels to produce an albedo kernel $K_a^{\texttt{cor}} = 0.5 (K_{a, \texttt{rgb}}^{\texttt{sym}} + K_{a, \texttt{xyz}}^{\texttt{cor}})$ which combines physical-space and RGB-space distance information.  By combining these albedo kernels with our shape kernel $K_a$, we can construct 3DMMs which possess a color-channel correlation heuristic but which lack an explicit symmetry heuristic.  We list these 3DMMs in Table~\ref{tab:correlated_kernels}.
    
    \begin{table}[hb]
        \centering
        \begin{tabular}{|l|c|c|}
            \hline
            name & shape kernel & albedo kernel\\
            \hline
            correlated-full & $K_s$ & $K_a^{\texttt{cor}}$\\
            \hline
            correlated-RGB & $K_s$ & $K_{a, \texttt{rgb}}^{\texttt{cor}}$\\
            \hline
            correlated-XYZ & $K_s$ & $K_{a, \texttt{xyz}}^{\texttt{cor}}$\\
            \hline
        \end{tabular}
        \caption{Our Gaussian processes for modeling faces with color-correlated but asymmetric kernels.}
        \label{tab:correlated_kernels}
    \end{table}
    
    \begin{table}
        \centering
        \begin{tabular}{|llll|} 
            \hline
            angle                               & $15^\circ$        & $30^\circ$        &  $45^\circ$ \\
            probe id                            & \texttt{140\_16}  & \texttt{130\_16}  & \texttt{080\_16} \\ 
            \hline
            standard-full                       & 84.7              & 69.9              & 54.2 \\
            standard-RGB                        & 76.3              & 57.8              & 28.9 \\
            standard-XYZ                        & 77.1              & 62.7              & 35.7 \\
            correlated-full                     & 92.4              & \textbf{87.1}     & 66.7 \\
            correlated-RGB                      & 71.5              & 58.6              & 37.8 \\
            correlated-XYZ                      & 76.7              & 61.4              & 45.8 \\
            symmetric-full                      & \textbf{93.2}     & 85.9              & \textbf{72.3} \\
            symmetric-RGB                       & 73.5              & 61.4              & 40.2 \\
            symmetric-XYZ                       & 73.1              & 58.6              & 44.2 \\
            \hline
        \end{tabular}
        \caption{Face recognition results on images from the Multi-PIE database \cite{gross_multi-pie_2010}.  Each column represents the accuracy for a set of probe images with a common yaw angle given in the first row.  The second row gives the common ending of the IDs in the Multi-PIE dataset of the probe images with a given yaw angle.  The gallery is constructed from images with a yaw angle of $0^{\circ}$ (dataset IDs ending in \texttt{051\_16}).}
        \label{tab:correlated_fr}
    \end{table}
    
    We repeat the face recognition experiment presented in Section 3.1.2 of our main paper with these 3DMMs (once again using the mean of the 2019 Basel Face Model \cite{gerig2018morphable} as our reference mesh).  The results of this experiment are shown in Table~\ref{tab:correlated_fr}, along with a copy of the results with the standard and symmetric 3DMMs that were shown in the main paper.  Table~\ref{tab:correlated_fr} demonstrates that the color-correlated asymmetric 3DMMs perform comparably to the symmetric (and color-correlated) 3DMMs on faces with $15^{\circ}$ and $30^{\circ}$ yaw angles.  On faces with $45^{\circ}$ yaw angles, they are significantly worse, indicating that (unsurprisingly) a symmetry prior becomes more important as the yaw angle increases.  Nevertheless, in general the color-correlated asymmetric 3DMMs perform quite well.  This indicates that in an inverse graphics context the color-channel correlation heuristic is more important to our symmetric 3DMMs than the symmetry heuristic is, at least for input images with a low yaw angle.

\section{Additional Bird and Fish Models}\label{sec:additional_birds_and_fish}

    \begin{figure}
        \begin{center}
            \includegraphics[width=0.9\linewidth]{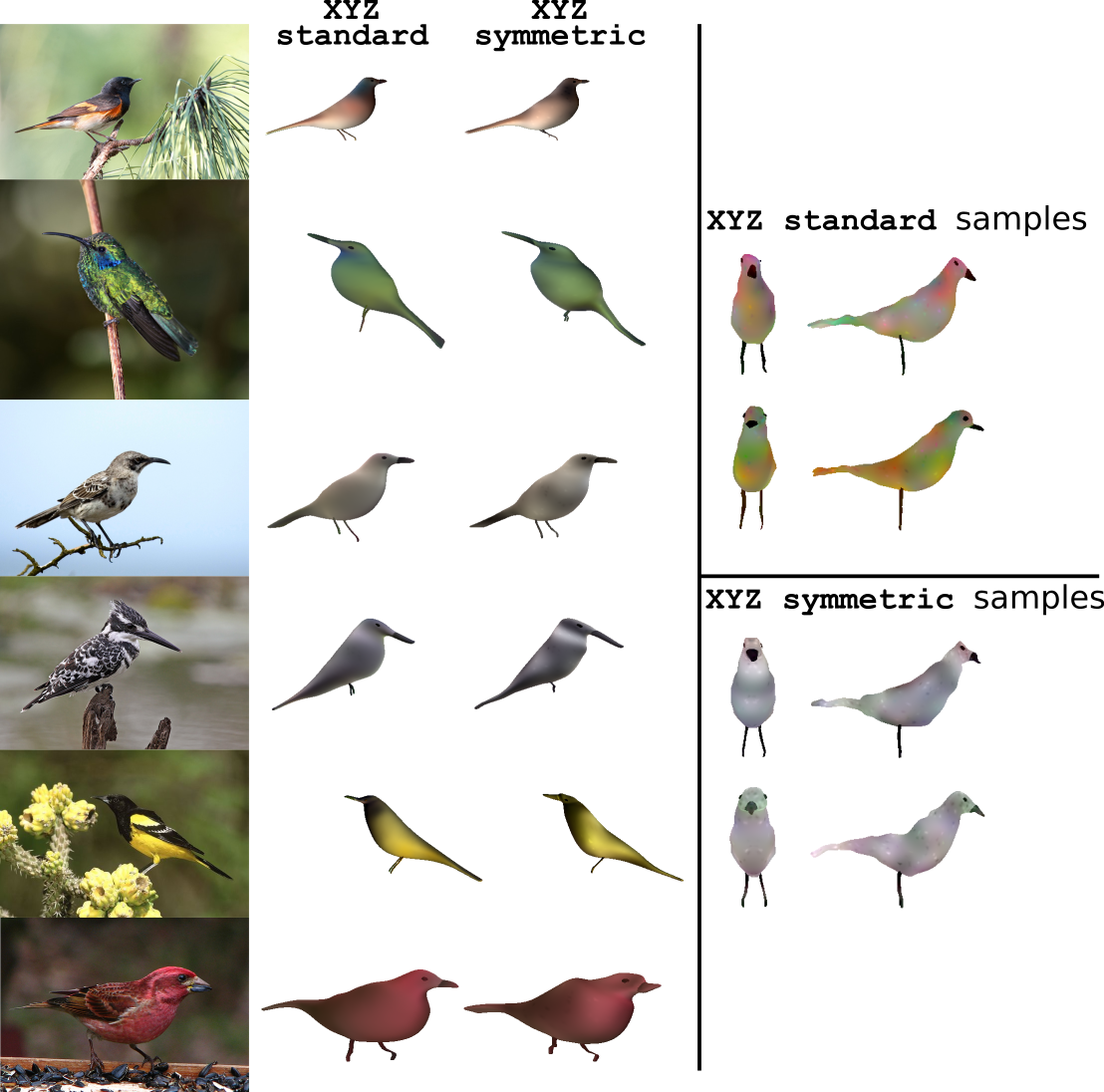}
        \end{center}
        \caption{On the left: the reconstructions produced by the two bird models built using only physical distance information on six images taken from the Caltech-UCSD Birds 200 dataset \cite{wah_caltech-ucsd_2011}.  On the right: samples from these models, shown in frontal and side views.}
        \label{fig:bird_lfw_2}
    \end{figure}

    \begin{figure}
        \begin{center}
            \includegraphics[width=0.9\linewidth]{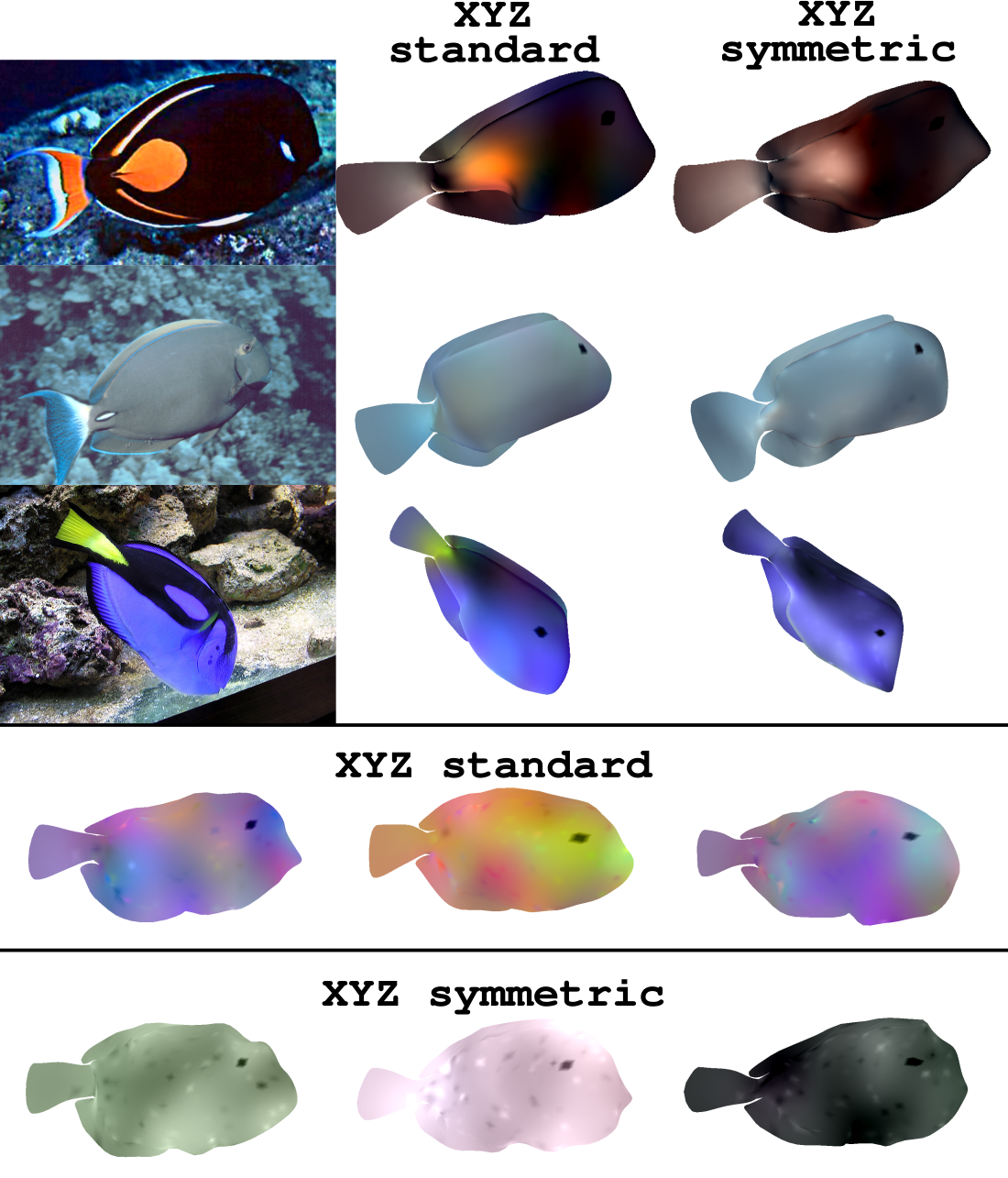}
        \end{center}
        \caption{On the top: the reconstructions produced by the two fish models built using only physical distance information on three natural images of fish.  On the bottom: samples from these models, shown in side views.}
        \label{fig:fish_lfw_2}
    \end{figure}

    In our paper we presented bird and fish 3DMMs created analogously to the standard-full and symmetric-full 3DMMs, i.e. with $K_s$ and $K_a$, and with $K_s^{\texttt{sym}}$ and $K_a^{\texttt{sym}}$, respectively.  We can also define similar bird and fish 3DMMs using albedo kernels that only rely on physical distance; i.e., using $K_{a, \texttt{xyz}}$ and $K_{a, \texttt{xyz}}^{\texttt{sym}}$ instead of $K_a$ and $K_a^{\texttt{sym}}$.  This produces two new 3DMMs for each reference mesh, which for space reasons are listed as listed as ``XYZ standard'' and ``XYZ symmetric''.  Figure~\ref{fig:bird_lfw_2} shows samples from these two bird 3DMMs, as well as reconstructions produced with these models of the bird images that were used in the main paper.  Figure~\ref{fig:fish_lfw_2} shows analogous samples and reconstructions for the two new physical-distance-based fish 3DMMs.  The results are close to those produced in the main paper; this is unsurprising given that the 3D mesh used to build these 3DMMs does not include complex coloration, and instead has near-piecewise-constant albedo.
    
    In both Figure~\ref{fig:fish_lfw_2} and in the main paper, we obtain our input natural fish images from Wikipedia\footnote{The links are:\\ \href{https://en.wikipedia.org/wiki/File:Acanthurus\_achilles1.jpg}{https://en.wikipedia.org/wiki/File:Acanthurus\_achilles1.jpg}, \href{https://en.wikipedia.org/wiki/File:Acanthurus\_dussumieri.jpg}{https://en.wikipedia.org/wiki/File:Acanthurus\_dussumieri.jpg}, and \href{https://en.wikipedia.org/wiki/File:Paracanthurus\_hepatus\_(Regal\_Tang).jpg}{https://en.wikipedia.org/wiki/File:Paracanthurus\_hepatus\_(Regal\_Tang).jpg}}.

\section{Additional Registration Results}\label{sec:additional_registration}
    
\begin{figure}
        \begin{center}
            \includegraphics[width=0.65\linewidth]{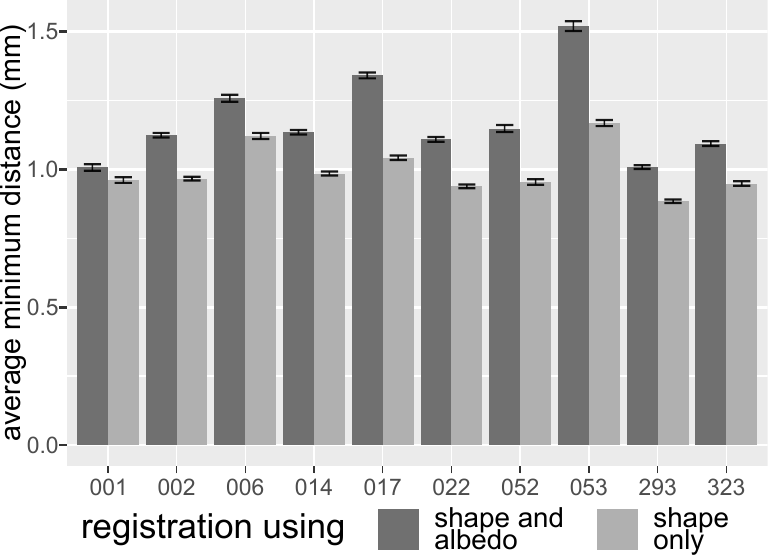}
        \end{center}
        \caption{The average distance between each vertex in each of the registered meshes and the closest point in the corresponding face scan, with error bars ($\pm 1.96$ standard error).}
        \label{fig:registration_chamfer_all}
    \end{figure}

\begin{figure}[h]
        \begin{center}
            \includegraphics[width=0.65\linewidth]{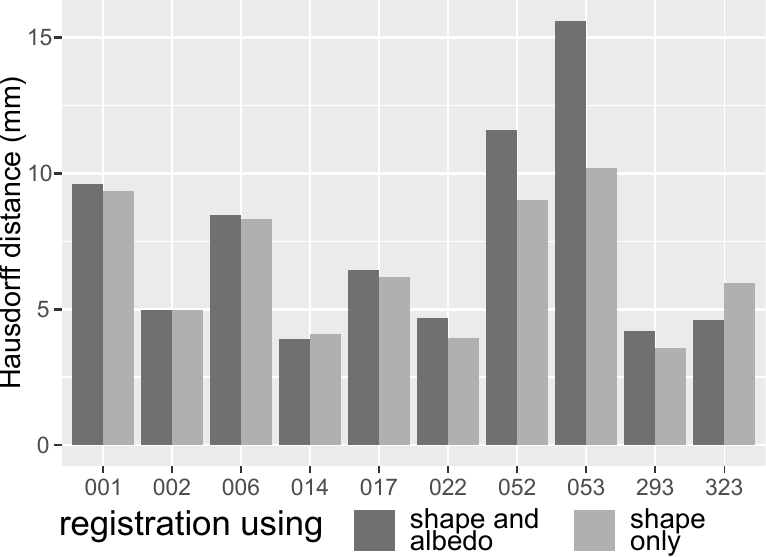}
        \end{center}
        \caption{The Hausdorff distance between the vertices of each of the registered meshes and the vertices of the corresponding face scan.}
        \label{fig:registration_hausdorff_all}
    \end{figure}

\begin{figure}
        \begin{center}
            \includegraphics[width=0.65\linewidth]{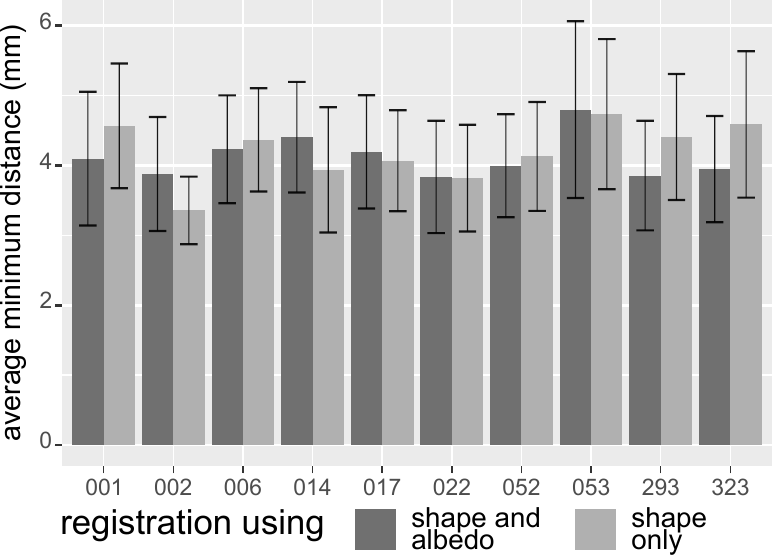}
        \end{center}
        \caption{The average distance between each landmark in each of the registered meshes and the closest landmark in the corresponding face scan, with error bars ($\pm 1.96$ standard error).}
        \label{fig:registration_chamfer_landmark}
    \end{figure}

\begin{figure}
        \begin{center}
            \includegraphics[width=0.65\linewidth]{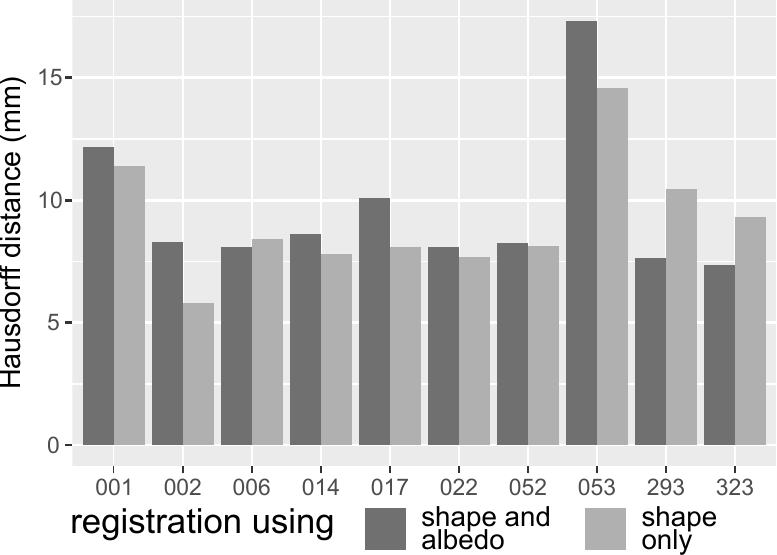}
        \end{center}
        \caption{The Hausdorff distance between the landmarks of each of the registered meshes and the landmarks of the corresponding face scan.}
        \label{fig:registration_hausdorff_landmark}
    \end{figure}

    Figures ~\ref{fig:registration_hausdorff_all}, ~\ref{fig:registration_chamfer_all}, ~\ref{fig:registration_hausdorff_landmark}, and ~\ref{fig:registration_chamfer_landmark} offer a variety of quantitative metrics of the shape error of the registered meshes produced in our paper's registration tasks.  As in the main paper, in all figures the ``shape and albedo'' option refers to meshes registered using both shape and albedo information in the MCMC method, while the ``shape only'' option refers to meshes registered using only shape information in the MCMC method.  These figures do not take into account the stability of the reconstruction or the albedo error.
    
    We estimate shape error through Hausdorff distance (Figures ~\ref{fig:registration_hausdorff_all} and ~\ref{fig:registration_hausdorff_landmark}) and chamfer distance (Figures ~\ref{fig:registration_chamfer_all} and ~\ref{fig:registration_chamfer_landmark}) between either the vertices of the registered meshes and the corresponding scans (Figures ~\ref{fig:registration_hausdorff_all} and ~\ref{fig:registration_chamfer_all}) or a sparse set of landmarks on the registered meshes and corresponding scans (Figures ~\ref{fig:registration_hausdorff_landmark} and ~\ref{fig:registration_chamfer_landmark}).  We obtained landmark information by using the landmark annotations given in \cite{paysan20093d} for each of the 10 input meshes and the landmark annotations provided with the 2019 Basel Face Model \cite{gerig2018morphable} for the registered meshes (since these have the same topology as the 2019 Basel Face Model).
    
    Figures ~\ref{fig:registration_hausdorff_all} and ~\ref{fig:registration_chamfer_all} demonstrate that including albedo information along with shape information slightly increases the shape reconstruction error.  As noted in the main text, this is to be expected; the shape-only reconstruction is optimized to produce the lowest shape error possible, whereas the reconstruction produced using both shape and albedo is also optimized to produce a low albedo error, and by definition cannot have a lower shape error than the reconstruction with the minimum shape error.  However, the increase in shape error is not very large.  Furthermore, Figures ~\ref{fig:registration_hausdorff_landmark} and ~\ref{fig:registration_chamfer_landmark} demonstrate that including albedo in registration does not significantly affect the shape error of landmarks.  This suggests that the incorporation of albedo information does not reduce the registration quality of the \textit{important aspects} of face shape.

\section{Qualitative Reconstructions}\label{sec:qualitative_lfw}

\begin{figure*}
        \begin{center}
            \includegraphics[width=0.95\linewidth]{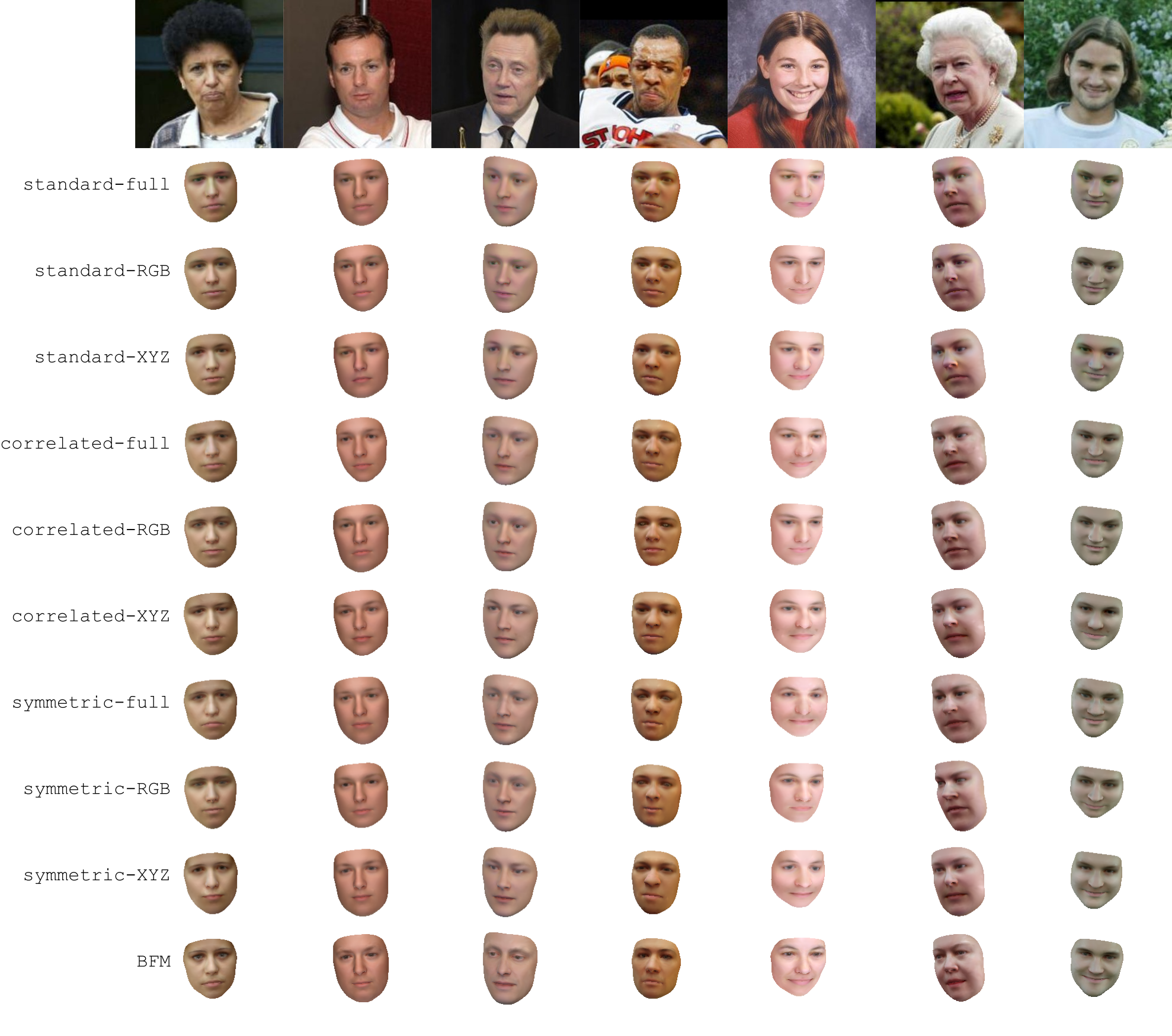}
        \end{center}
        \caption{The face reconstructions produced from all the 3DMMs built using the mean of the 2019 Basel Face Model \cite{gerig2018morphable} on natural images from the Labeled Faces in the Wild dataset \cite{huang2008labeled}, as well as the reconstructions produced using the 2019 Basel Face Model itself (``BFM'').}
        \label{fig:lfw_results_full}
    \end{figure*}

\begin{figure*}
        \begin{center}
            \includegraphics[width=0.80\linewidth]{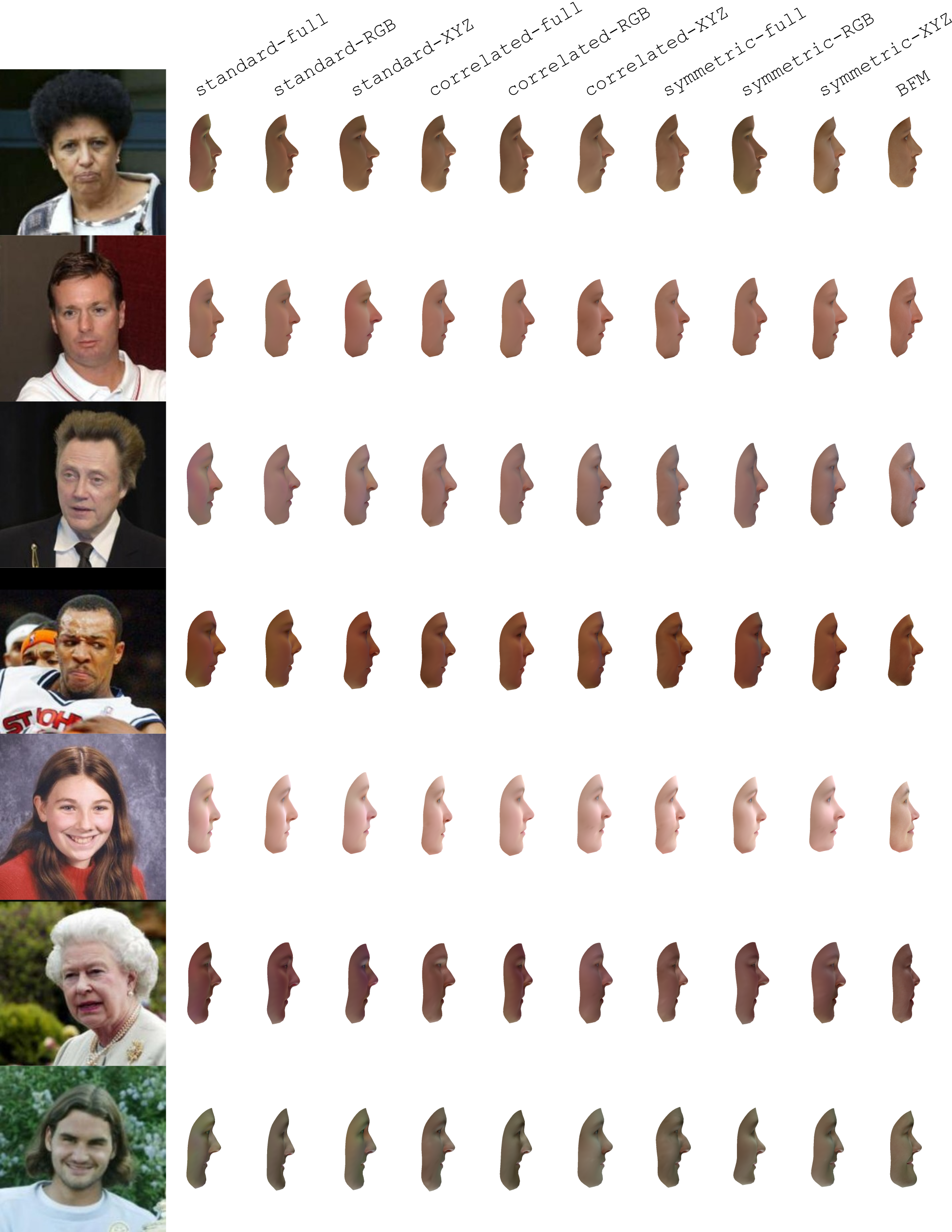}
        \end{center}
        \caption{Side views of the reconstructions presented in Figure~\ref{fig:lfw_results_full}.}
        \label{fig:lfw_results_side_view}
    \end{figure*}

\begin{figure*}
        \begin{center}
            \includegraphics[width=0.95\linewidth]{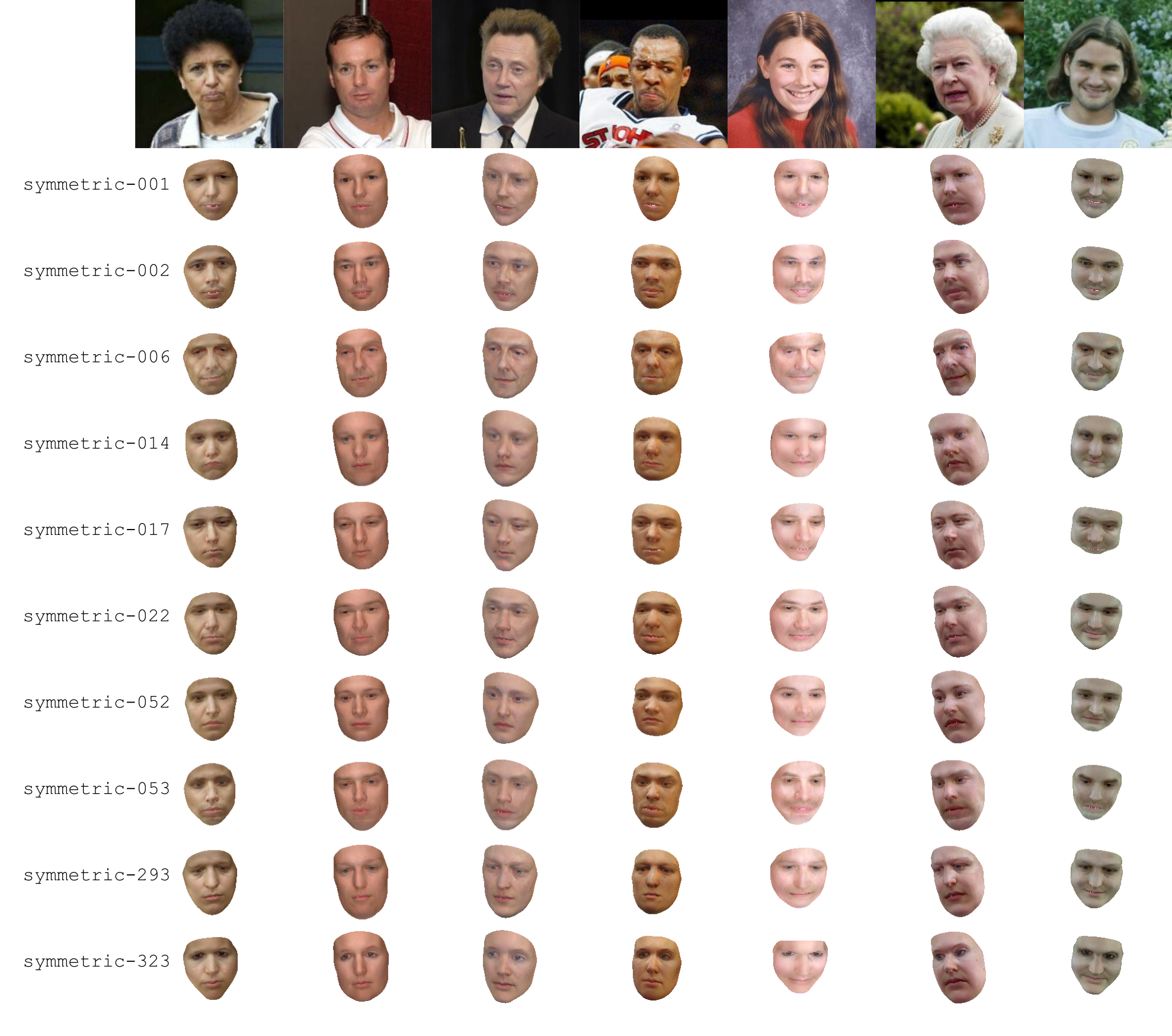}
        \end{center}
        \caption{The face reconstructions produced from all 3DMMs built using the symmetric-full kernel and scans included with the 2009 Basel Face Model \cite{paysan20093d} on natural images from the Labeled Faces in the Wild dataset \cite{huang2008labeled}.}
        \label{fig:lfw_results_non_mean}
    \end{figure*}

\begin{figure*}
        \begin{center}
            \includegraphics[width=0.95\linewidth]{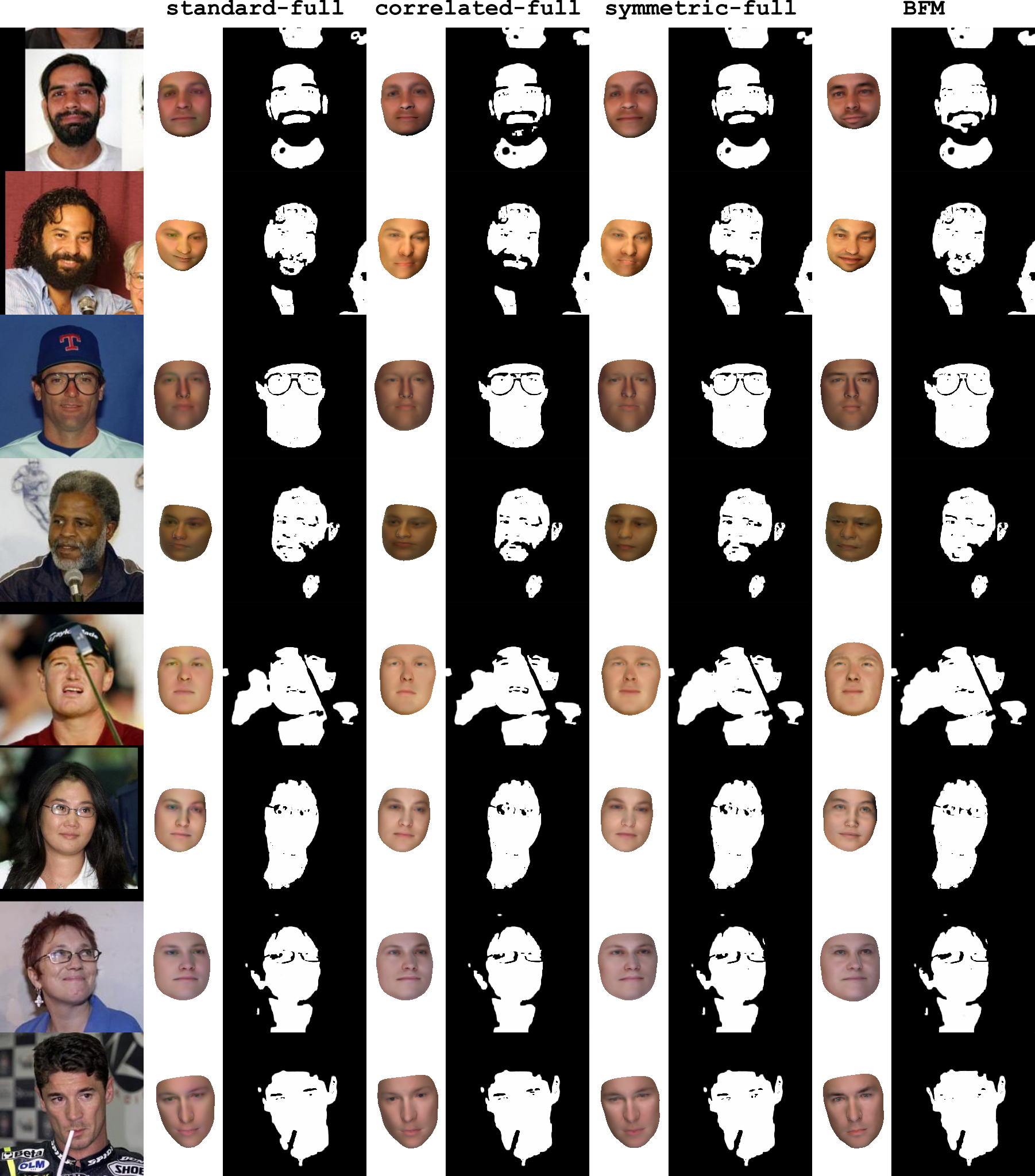}
        \end{center}
        \caption{The face reconstructions produced by our standard-full, correlated-full, and symmetric-full models, as well as the 2019 Basel Face Model \cite{gerig2018morphable} (``BFM''), on images from the Labeled Faces in the Wild dataset \cite{huang2008labeled}, produced using the occlusion-aware MCMC method described in \cite{egger2018occlusion}.  Both the segmentation masks and face reconstructions were inferred purely with top-down inference.}
        \label{fig:lfw_results_occluded}
    \end{figure*}

    Figures ~\ref{fig:lfw_results_full}, ~\ref{fig:lfw_results_side_view}, ~\ref{fig:lfw_results_non_mean}, and ~\ref{fig:lfw_results_occluded} provide additional qualitative reconstruction results.  Figures ~\ref{fig:lfw_results_full} and ~\ref{fig:lfw_results_side_view} present qualitative reconstructions (in frontal and side views, respectively) of images from the Labeled Faces in the Wild dataset \cite{huang2008labeled} produced using all the 3DMMs constructed using the mean of the 2019 Basel Face Model \cite{gerig2018morphable}.   Figure~\ref{fig:lfw_results_non_mean} presents qualitative reconstructions of the same images produced using 3DMMs built from the scans included with the 2009 Basel Face Model \cite{paysan20093d}.  These reconstructions are significantly lower-quality, because a significant portion of the shape of the template mesh is preserved during the MCMC process.  Figure~\ref{fig:lfw_results_occluded} presents qualitative reconstructions of different images from the Labeled Faces in the Wild dataset that contain significant occlusion.  Figure~\ref{fig:lfw_results_occluded} includes both 3D reconstructions as well inferred occlusion masks.  Figure~\ref{fig:lfw_results_occluded}'s results were produced using the occlusion-aware MCMC method described in \cite{egger2018occlusion}.
    
\clearpage
\end{appendices}

\end{document}